\documentclass[10pt]{article} 

\usepackage[preprint]{tmlr}

\usepackage{amsmath,amsfonts,bm}

\def\eqref#1{equation~\ref{#1}}

\def\1{\bm{1}}

\DeclareMathAlphabet{\mathsfit}{\encodingdefault}{\sfdefault}{m}{sl}
\SetMathAlphabet{\mathsfit}{bold}{\encodingdefault}{\sfdefault}{bx}{n}

\newcommand{\E}{\mathbb{E}}

\DeclareMathOperator*{\argmax}{arg\,max}
\DeclareMathOperator*{\argmin}{arg\,min}

\usepackage{hyperref}
\usepackage{url}
\usepackage{svg}
\usepackage{graphicx}

\title{Probabilistic 3d regression with projected huber distribution}

\author{\name David Mohlin \email davmo@kth.se \\
      \addr RPL,KTH/Tobii
      \AND
      \name Josephine Sullivan \email sullivan@kth.se\\
      \addr RPL,KTH
      }

\def\month{MM}  
\def\year{YYYY} 
\def\openreview{\url{https://openreview.net/forum?id=XXXX}} 
\newcommand{\OurDist}{\textit{Projected Huber Distribution }}

\begin{document}

\maketitle

\begin{abstract}
Estimating probability distributions which describe where an object is likely to be from camera data is a task with many applications. In this work we describe properties which we argue such methods should conform to. We also design a method which conform to these properties. In our experiments we show that our method produces uncertainties which correlate well with empirical errors. We also show that the mode of the predicted distribution outperform our regression baselines. The code for our implementation is available \href{https://github.com/Davmo049/public\_probabilistic\_3d\_regression}{online}
\end{abstract}

\begin{figure}[h]
\includegraphics[width=\linewidth]{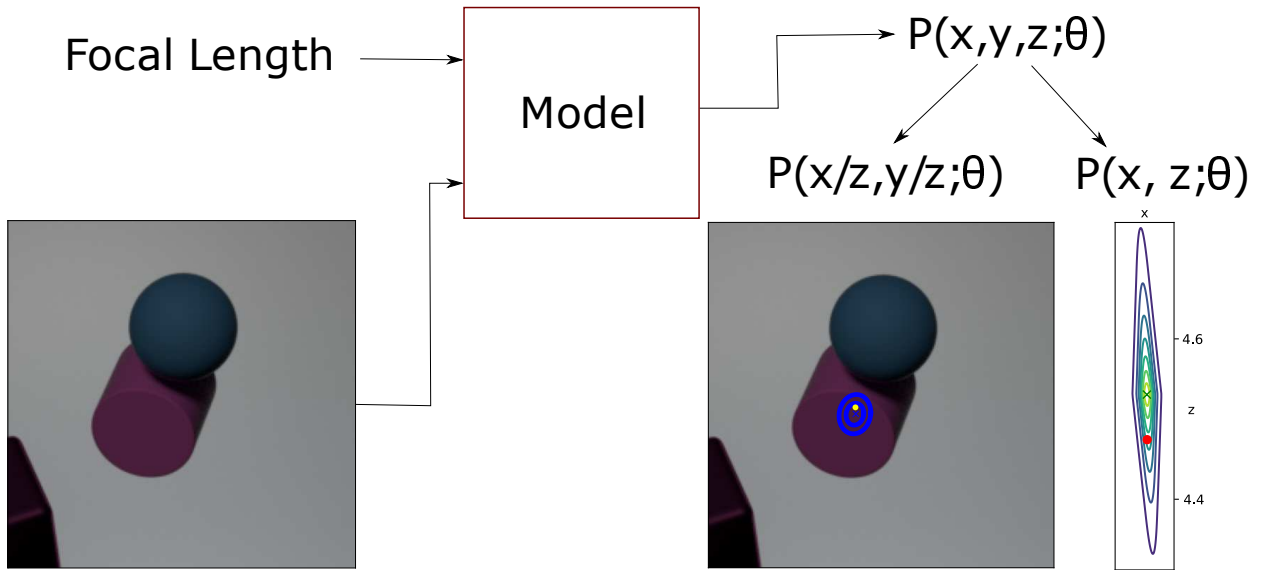}
\caption{Overview of out method. Our method takes an image as input and the focal length which was used to capture the scene and produces a log concave probability distribution in world coordinates in a way which can model the ambiguities which are inherent for camera sensors.
On the left we have visualized the level curves in projected coordinates with the projected ground truth center of the cylinder. On the right we show the level curves of the distribution from a birds eye view. The estimated mode is shown as a cross and the ground truth position is a dot.}
\label{fig:splash}
\end{figure}

\section{Introduction}
\label{sec:introduction}
Estimating 3d position of object have many use cases.
For example 1) Estimating the position of a car or pedestrian is required to construct policies for how to traverse traffic for autonomous driving or ADAS. As a result this task is included in many such datasets, for example KITTI \cite{Geiger2012CVPR}.
2) Body pose estimation can be seen as trying to find the 3d position of each joint for a human skeleton. Examples of such datasets are \cite{vonMarcard2018} and \cite{ionescu2013human3}.
3) For many robotics applications estimating the 3d position of objects is relevant \cite{bruns2022evaluation}. For industrial robotics the shape of the object is often known, therefore the only source of variation is the six degrees of freedom due to the orientation and position of the object.

Estimating uncertainties associated with the estimated position has applications for fusion of multiple estimates  \cite{mohlin2021probabilistic}. Another application for uncertainty estimates is time filters, such as Kalman filters. Uncertainties can also be used for application specific purposes. For examples in an ADAS setting it makes sense to not only avoid the most likely position of an object, but to also include a margin proportional to the uncertainty of the estimate. In robotic grasping a good policy could be to try to pick up objects directly if the estimated position is certain while doing something else if the estimate is uncertain, such as changing viewing angle.

\section{Prior work}
\label{sec:prior_work}
When estimating 3d position from camera data there is an inherent ambiguity due to scale/depth ambiguity. Many types of interesting types of objects have a scale ambiguity, such as people, cars and animals.

The existence of scale ambiguity is reflected in the performance on competitive datasets. For example the errors of single view methods which estimate the position of people relative to the camera is on the order of 120mm where 100mm is in the depth direction \cite{moon2019camera}. This is consistent with the presence of scale/depth ambiguity. In the multi view setting where the depth ambiguity can be eliminated the state of the art is on the order of 17mm \cite{iskakov2019learnable}.

There are many ways to resolve the scale/depth ambiguity.

One way to solve the problem is to add a depth sensor such as a lidar \cite{Geiger2012CVPR} or structured light \cite{choi2016large}. However adding additional sensors come with several drawbacks such as price, complexity, range among others.
For these reasons we will focus on methods which only use camera data.

For camera only methods there are two common ways to resolve the ambiguity, the first is only applicable if the object lies on a known plane, in practice this plane is often the ground plane, but could also be for example the surface of a table. \cite{mills1989force}
Another approach is to do multi-view fusion. In this case the errors due to these ambiguities point in different directions and can therefore be cancelled out with a suitable method.

For this reason it is important if a method to estimate 3d position also fits into a framework which is able to resolve the scale/depth ambiguity.

It is well known that using densities which are log concave are well suited for sensor fusion since computing the optimal combination is a convex problem in this setting \cite{an1997log}. In our work we show that both the ground plane assumption and multi view fusion turns into a convex optimization problems if model predicts a log concave probability distribution.

There are many other works which treat the problem of estimating 3d position in a probabilistic framework.
For example \cite {feng2019can} models the the position of an object with a normal distribution. However they also use a detector on lidar data to avoid large errors and thereby large gradients.
\cite{meyer2019lasernet} models the locations of the corners of a bounding box with a laplace distributions, but they also avoid the scale/depth ambiguity by using lidar data.
Many works treat depth estimation as a classification problem were a final prediction is constructed by computing the expected value of this discrete distribution \cite{wang2022probabilistic}, but the probabilities predicted by such methods are in general not concave since it is possible to model a multimodal distribution with these methods.
In \cite{Bertoni_2019_ICCV} they model the reciprocal of the depth with a laplace distribution. However they do not model the projected location in a probabilistic manner, which is not well suited for multi-view fusion. In section \ref{subsec:properties_given_constraints} we also show that modelling depth independently of the projected location always result in undesirable properties, which they, like many others do.

Some works investigate regressing 2d position by having neural networks estimate probability distributions parameterized by mean and covariance \cite{kumar2020luvli, mohlin2021probabilistic}

\section{Motivation}
\label{sec:motivation}
This section will focus on the specific properties which data captured by cameras conform to and therefore which a model should take into account when doing probabilistic position estimation based on camera data.

In practice when applying this method the location of the object is modeled by a probability distribution which is parameterized by the output of a convolutional neural network.

\subsection{Notation}
We will investigate a setting where the model takes an input image which is produced by capturing a scene on a $S\times S$ sensor.
The camera is a pin-hole camera without radial distoritions and have the known intrinsics
\begin{equation}
\begin{bmatrix}
f & 0 & S/2 \\
0 & f & S/2 \\
0 & 0 & 1 \\
\end{bmatrix}
\end{equation}
Coordinates for this camera are denoted $v = (x,y,z)$ where the $z$ axis is aligned with the principal axis of the camera and $v=\bar{0}$ correspond to the center of the camera.

The method will predict a probability distribution parameterized by $\theta$ based on the input image.

\subsection{Desired constraints for estimated distribution}
In this section we describe constraints which we argue a probability distribution estimated from camera data should conform to. We also motivate why these constaints should exist.

\textbf{constraint 1}:
The model can express any variance for the projected coordinates on the image sensor and depth independently.

Formally:

\begin{equation}
\forall (a,A) \in \mathbb{R}^{+}\times\textbf{S}^2_{++} \exists \theta \text{ such that } Var[z|\theta] = a \text{ and } Var[x/z, y/z|\theta] = A
\end{equation}

Motivation

For many tasks the error of depth estimation is inherently large due to scale/depth ambiguity. Therefore for objects which have a variation in size, but where the absolute size is hard to estimate there will be an inherent error proportional to the scale error and the distance to the object. Despite this estimating the position in projected coordinates can often be done accurately.
For example assuming that the height of humans is between 1.5-1.9m and assuming that it is intrinsically hard to estimate the size of an unseen person an uncertainty of 13\% of the distance to the person is reasonable to expect. 
Despite this it is possible to estimate the position of keypoints in the projected coordinate space of an image accurately, often within a few pixels. \\

\textbf{constraint 2: } The model should not try to estimate the coordinates directly, but instead estimate the depth divided by the focal length and the projected position of the object.

Formally:

\begin{equation}
p_{cam}(x,y,z|\theta(image); f) = p_{cam}(fx/z,fy/z,z/f|\theta(image))
\end{equation}

Motivation:

It is difficult to separate the the quantities scale, depth and focal length, since the main effect of each variable is to change the apparent size of the object on the sensor. For this reason the focal length needs to be taken into account when estimating the depth of an object, otherwise the variation in focal length adds additional ambiguity.
When scale/depth ambiguity is present it is not possible to estimate the absolute position for the x or y coordinate accurately, but it is often possible to do so for the projected position of the object. This motivates why a method should also estimate the location in image space. Methods which conform to these constrants are not uncommon. For example \cite{zhen2020smap} estimates the depth after dividing by the focal length. It is also common to estimate the location of objects in image space.

\textbf{constraint 3:} The output probability should have support only for points in front of the camera.

Formally:
\begin{equation}
p_{cam}(x,y,z) = 0 \text{ if } z < 0
\end{equation}

Motivation:

Cameras only depict objects in front of them. The output distribution should reflect that.

\textbf{constraint 4:} The output probability should have support for all points in the field of view.

Formally:
for a camera with focal length $f$ and sensor size $S$
\begin{equation}
\forall x,y,z \text{ such that } |fx/z| < S/2, |fy/z| < S/2 \text{ and }z>0 \text{ then } p_{cam}(x,y,z|\theta) > 0
\end{equation}

Motivation:

One standard way to optimize neural networks which output probability distributions is to minimize the negative log likelihood. Backpropagation is only possible if the predicted probability is larger than 0 for the ground truth position. Since there are not any guarantees what an untrained network predicts the probability has to be positive for all locations which are reasonable without considering the input data. Since a camera only depict objects in its field of view this is a sufficient condition.

\textbf{constraint 5:} The negative log likelihood of the position is convex.
Formally:
\begin{equation}
-\log(p_{cam}(v|\theta))
\end{equation}
is convex with respect to $v$

Motivation:

A common method to combine estimates which are expressed as probability distributions is by assuming that the two estimates have errors which are independent random variables and letting the output of the fusion be the maximum likelihood point under this assumption \cite{mohlin2021probabilistic}.
It is desirable if computing this combination is a convex optimization problem since it guarantees that it is possible to find the fusion quickly.
In section \ref{subsec:properties_given_constraints} we show that a sufficient constraint for this property is that the negative log likelihood is convex with respect to position.

\textbf{constraint 6:} The probability for objects infinitely far away from the camera is 0
Formally
\begin{equation}
\lim\limits_{r \rightarrow \infty} \max\limits_{\|v\| = r} p(v) = 0
\end{equation}
\begin{equation}
\forall x,y \in \mathbb{R}^2 \text{  } p(x,y,0) = 0
\end{equation}

Motivation:

Firstly objects of finite size are not visible if they are infinitely far away from the camera. Secondly we need this property to guarantee convergence when doing multi-view fusion.

\textbf{constraint 7:} 
$p_{cam}(v|\theta) $ is continuous with respect to $v$.

Motivation:

It is not reasonable that small changes in position change the density by a large amount.

\subsection{Properties given these constraints}
\label{subsec:properties_given_constraints}

\textbf{Proposition 1:}
All distributions which are twice differentiable at least at one point and decompose into
\begin{equation}
    p_{composed}(x,y,z|\theta) = p_{projected}(x/z, y/z|\theta)p_{depth}(z|\theta)
\end{equation}
Do not fulfill all of the above constraints.
Proof in supplementary \ref{sec:supp_limited_precision_for_decompose_vpz}

\textbf{Proposition: 2} In multi view fusion, if the field of view for the cameras have a non-empty intersection then if each camera produce probability estimates conforming to constraints 1-7 then a valid maximum likelihood fusion point exist and can be found by convex optimization. Proof in supplementary \ref{sec:supp_multi_view_ground_plane}

\textbf{Proposition 3:}
Imposing a ground plane constraint is a convex optimization problem. If any point of the ground plane is in the field of view of the camera, then the constraints also guarantee that a solution exist. Proof in supplementary \ref{sec:supp_multi_view_ground_plane}

\section{Method}
\label{sec:method}
In this work we first describe the \OurDist which we will show fulfills the constraints described in section \ref{sec:motivation}

\subsection{\textbf{\OurDist}}

The distribution can be decomposed into one component $p_{proj}$ which mainly model the probability in the projected coordinates and another component $p_{depth}$which models the distribution in depth

The component which mainly models the probability over the projected coordinates is

\begin{equation}
    p_{proj}(x,y |z;\mu, A) = \dfrac {1}{K_{depth}(A,\mu_z)} \exp\left(-h\left(\left\|A\begin{bmatrix} x/z-\mu_x \\ y/z-\mu_y \end{bmatrix}\right\|_2\dfrac{z}{\mu_z}\right)\right)
\end{equation}

By including $z$ in this distribution we can avoid the problem described in proposition 1. Note that conditioning on $z$ is necessary to get a proper distribution.

$\mu_x$, $\mu_y$ $\in \mathbb{R}^2$ model the mean position on the camera sensor. $\mu_z \in \mathbb{R}^{+}$ models the estimated depth of the object and $A \in \textbf{S}^2_{++}$ models the precision. The function $h(.)$ is a huber function.

The distribution which models the depth is

\begin{equation}
p_{depth}(z;\mu_z, a) = \begin{cases} \dfrac {1}{K_{depth}(a, \mu_z)} \exp(-a\max(z/\mu_z, \mu_z/z)) & \text{ if } z > 0 \\
0 &\text{ otherwise} 
\end{cases}
\end{equation}

Where $\mu_z \in \mathbb{R}^{+}$ models the estimated depth. Note that this is the same parameter as for $p_{proj}$.
$a \in \mathbb{R}^{+}$ roughly models the precision, that is a larger $a$ reduces the variance.

By combining these distributions we get the \OurDist
\begin{align}
    \label{eq:our_prob}
    p_{combined}(x,y,z; \mu, A,a) = p_{proj}(x,y|z; A, \mu)p_{depth}(z;\mu_z, a)
\end{align}

for x,y,z in the cameras coordinate system. 

The normalizing factors are 
\begin{align}
K_{depth}(\mu_z, a) &= \mu_z (\exp(-a)/a+\Gamma(-1,a)a) \\
K_{proj}(\mu_z,A) &= \dfrac {\mu_z^2} {|A|}2\pi(1+\exp(-1/2)) \\
K_{combined}(\mu_z, A,a) &= K_{depth}(\mu_z, a)K_{proj}(\mu_z,A)
\end{align}

Where $K_{depth}$ is bounded by
\begin{equation}
    \mu_z\dfrac {\exp(-a)}{a} \le K_{depth}(\mu_z, a) \le \mu_z\exp(-a)(1/a+1)
\end{equation}

This distribution will conform to all constraints, except constraint 2 which relates to how to predict the parameters of the model. We show how to predict the parameters in a way which conform to constraint 2 in subsection \ref{subsec:parameter_remapping}

\textbf{proposition 5:}
the moments of this distribution are
\begin{align}
E\left[\begin{bmatrix} x/z \\ y/z \end{bmatrix}|\mu_p\right] &= \mu_p \\
\label{eq:analytical_var_projected}
Var[\begin{bmatrix} x/z \\ y/z \end{bmatrix}|A] &= A^{-2}\dfrac {4+3\exp(-1/2)} {2+2\exp(-1/2)} \\
E\left[z\right] &= \mu_z\dfrac{\Gamma(2,a)/a^2+\Gamma(-2,a)a^2}{\Gamma(1,a)/a+\Gamma(-1,a)a} \\
\label{eq:analytical_var_depth}
Var\left[z\right] &= \mu_z^2\dfrac{(\Gamma(3,a)/a^3+\Gamma(-3,a)a^3)(\Gamma(1,a)/a+\Gamma(-1,a)a)-(\Gamma(2,a)/a^2+\Gamma(-2,a)a^2)^2}{(\Gamma(1,a)/a+\Gamma(-1,a)a)^2}
\end{align}

The $\Gamma$ terms are not very intuitive, but the fraction for the expected value quickly decrease to 1 while the fraction for the variance behaves similar to $1/a^2$ that is

\begin{align}
    E\left[z\right] &\approx \mu_z \\
    Var\left[z\right] &\approx \mu_z^2/a^2
\end{align}

proof in supplementary \ref{sec:supp_moments}

\subsection{Proof for constraints}
\textbf{constraint 1:}
Proof: From proposition 5 we see that $A$ models the uncertainty in projected coordinates, while $a$ models the uncertainty in depth coordinates.

\textbf{constraint 3:}
$p(z) = 0$ if $z < 0$, from definition in equation \ref{eq:our_prob}

\textbf{constraint 4:}
$p(z) > 0$ if $z > 0$ since the range of $\exp$ is the $(0, \infty)$ and the expression can be evaluated for all values of $x,y,z \in \mathbb{R}^2\times\mathbb{R}^{+}$.
The field of view for a camera is a subset of the half plane $z>0$.

\textbf{constraint 5:}
Proof sketch:
Show that both $p_{proj}$ and $p_{depth}$ has a convex negative log likelihood. An affine basis change turns the argument of the $-\log(p_{proj})$ into a norm $h(\|q\|_2)$ which is convex. $-\log(p_{depth})$ is also convex.
Full proof in supplementary \ref{sec:supp_convexity}

\textbf{constraint 6:}
Proof sketch. For the region $z \le 0$ the proof is trivial.
Then we prove it for the region $0 < z <= (x^2+y^2)^{1/4}$. Here the $\exp(-h(.))$ term will go to 0. as $\|v\|_2 \rightarrow \infty$ while the other factors are bounded.
For the region $z > (x^2+y^2)^{1/4}$
the factor $\exp(-a\max(z/\mu_z, \mu_z/z))$ goes to 0 while the other factors are bounded. Full proof in supplementary \ref{sec:supp_proof_constraint_6}

\textbf{constraint 7:}
Proof sketch for $z > 0$ the function is trivially continuous. For $z < 0$ the function is also trivially continuous.
When z approaches 0 the factor $\exp(-a\max(z/\mu_z, \mu_z/z))$ goes to 0.
Full proof in \ref{supp:proof_continous}

\subsection{Parameter remapping}

When the parameters $\theta$ of \OurDist are estimated by a neural network distribution it is necessary to turn the output of the neural network into a valid parameterization for \OurDist.
Valid in this sense refers to conforming to the constraints $A \in \textbf{S}^{++}$, $a \in \mathbb{R}^{+}$ and $\mu_z \in \mathbb{R}^{+}$, that is $A$ is positive definite while a and $\mu_z$ are positive.

Neural networks give outputs in $\mathbb{R}^{d_{out}}$ where $d_{out}$ is decided by the architecture of the network. 
These outputs are do not conform to our desired parameter constraints, unless a suitable activation is applied.

To construct this activation we start by doing a basis change to conform to \textbf{constraint 2} from the motivation.

\begin{align}
\label{eqs:basis_change_variables}
    v_p &= \begin{bmatrix} x_p \\ y_p \end{bmatrix} = \begin{bmatrix} \dfrac{2fx} {zS} \\ \dfrac{2fy} {zS} \end{bmatrix} \\  
     z_p &= \dfrac{z} {\mu_{z0}f}
\end{align}

Where $\mu_{z0}$ is a bias term defined by

\begin{equation}
\label{eq:muz0}
    \mu_{z0} = \sqrt{\left(\max\limits_{z,f \in \textit{Dataset}}z/f\right)\left(\min\limits_{z,f \in \textit{Dataset}}z/f\right)}
\end{equation}

For the purpose of proving that our loss has bounded gradients we also need the constant $D$ defined by

\begin{equation}
\label{eq:D}
    D = \sqrt{\left(\max\limits_{z,f \in \textit{Dataset}} z/f \right)\bigg/ \left(\min\limits_{z,f \in \textit{Dataset}}z/f\right)}
\end{equation}

These values can either be derived from known properties of the dataset or by computing these values for all samples in the training dataset.

$f/z$ proportional to the projected scale of an object. Even in the extreme case where the scale ambiguity is 40\% and the projected size varies between 2 and 200 pixels $D$ would be less than 20. Furthermore $D$ is only used to prove that the loss has bounded gradients.

\textbf{proposition 5}
$\|v_p\|_{\infty} \le 1$ and $1/D \le z_p\ \le D$.
for points in the field of the camera with a depth conforming to equation \ref{eq:D}.

\textbf{proof}
Proof sketch $v_p$ correspond to the projected image coordinates, scaled to be between -1 and 1. $z_p$ is normalized by defininition \ref{eq:muz0}-\ref{eq:D}.
Full proof in supplementary \ref{supp:proof_bound_gt_variables}

Now $v_p$ correspond to the coordinates in the projected image and $z_p$ is the z coordinate after compensating for the scale change of the focal length and applying a logarithm to map $\mathbb{R}^{+}$ to $\mathbb{R}$

Define
\begin{align}
\label{equation:basis_change_parameters}
    \nu_p &= \begin{bmatrix} \nu_x \\ \nu_y \end{bmatrix} = \dfrac {f\mu_{z0}} {\mu_z} A \begin{bmatrix} \mu_x \\ \mu_y \end{bmatrix} \\
    \nu_z &= \dfrac {\mu_z} {\mu_{z0}f}\\
    B &= A\dfrac {Sf\mu_{z0}} {2\mu_z}
\end{align}

Using a negative log likelihood of $p_{combined}$ in the original basis can be written as 
\begin{align}
L_{combined\_original}(\mu, A, a; x,y,z) &= L_{proj\_original}(\mu_z, a, z) + L_{depth\_original}(\mu, A; x,y,z) \\
L_{proj\_original} &= h\left(\left\|A\begin{bmatrix} x/z-\mu_x \\ y/z-\mu_y \end{bmatrix}\right\|_2\dfrac{z}{\mu_z}\right) - \log(|A|) + 2\log(\mu_z) \\
L_{depth\_original} &= a\max(z/\mu_z, \mu_z/z) + \log(\exp(-a)/a+\Gamma(-1,a)a) +\log(\mu_z)
\end{align}

Which in the new basis is

\begin{align}
    \label{eq:loss_after_basis_change}
    L_{combined}(\nu, B, a; v_p, z_p) &= L_{proj}(\nu_p, A, v_p,z_p) + L_{depth}(\nu_z, a; z_p) \\
L_{proj}(B,\nu_p; v_p, z_p) &= h\left(\left\|Bv_p-\nu_p\right\|_2z_p)\right) - \log(|B|) \\
L_{depth}(a,\nu_z; z_p) &= a\max(z_p/\nu_z, \nu_z/z_p) + \log(\exp(-a)/a+\Gamma(-1,a)a) + \log(\nu_z)
\end{align}

In these losses we exclude terms not required for computing gradients such as terms only containing constants, $f, S, \mu_{z0}$.

$K_{depth}(a)$ and its gradients is computed by numerical integrals. The full method is described in supplementary \ref{sec:supp_normalizing_factor}

Estimating these parameters will conform to \textbf{constraint 2} since $\nu_p$ and $B$ models the position and precision for the projected coordinates while $\nu_z$ and $a$ models the depth and precision for the depth estimate.

\subsection{Enforcing constraints on distribution parameters}
\label{subsec:parameter_remapping}

Designing the activation which outputs $B$ and $\nu_p$ is done in the same way as \cite{mohlin2021probabilistic}.

Estimating $a$ and $\nu_z$ is done by starting with two real numbers $w_1$, $w_2$
\begin{equation}
    a(w_1) = \begin{cases}
    w_1+1 &\text{ if } w_1 < 0 \\
    exp(w_1) &\text{ otherwise}
    \end{cases}
\end{equation}
\begin{equation}
    \nu_z(w_1, w_2) = \begin{cases}
       1+w_2/a(w_1) &\text {if } w_2 > 0 \\
       1/(1- w_2/a(w_1)) &\text {otherwise}
    \end{cases}
\end{equation}

With this parameterization the the loss is convex with respect to the network output when $a>1$. The gradients of the loss will be bounded with respect to $w$. Proof in \ref{supp:convex_parameters} and \ref{supp:bound_grad}

Having a loss which is Lipschitz-continous should aid with stability during training since it avoids back-propagating very large gradients.

\section{Experiments}
\label{sec:experiments}
\begin{figure}[ht]
\includegraphics[width=\linewidth]{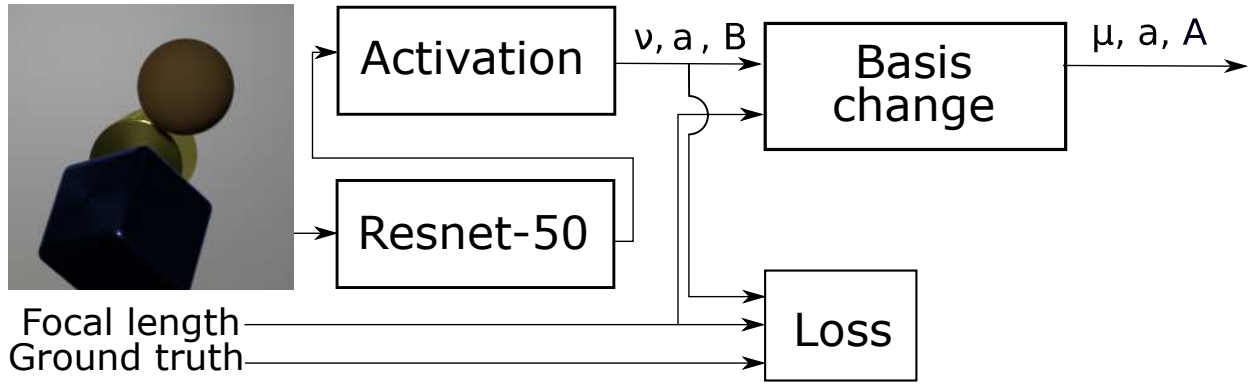}
\caption{Setup for experiments, images are fed into a standard resnet-50, the output of which is sent through the activation, basis change and loss described in section \ref{subsec:parameter_remapping}
}
\label{fig:experimental_setup}
\end{figure}

In this section we show how the method described in section \ref{sec:method} can be applied in practice to estimate the position of an object.
\subsection{Dataset}
We construct a synthetic dataset by rendering objects in a similar way as \cite{johnson2017clevr}. We choose our task to be to estimate the location of a rendered cylinder. 
The cylinder is rendered different reflection material properties, color and with random orientation.

The scene is rendered on a sensor of size $224 \times 224$ pixels, the focal length of the camera is sampled from a uniform distribution $f \sim U(1200, 2000) (pixels/m)$ The cylinder has a height of 0.2 meters and a radius of 0.1 meter. The depth is uniformly sampled from $z \sim U(3, 5)m$ The x and y coordinates are sampled from $\dfrac {Sz} {f} U(-0.5, 0.5)$ for each axis.

In this experimental setup the normalizing constants are
\begin{align}
\mu_{z0} &= \sqrt{\left(\max\limits_{f \in (1200,2000), z \in (3, 5)} z/f\right) \left(\min\limits_{f \in (1200,2000), z \in (3, 5)} z/f\right)} \approx 2.5*10^{-3} \\
D &= \sqrt{\left(\max\limits_{f \in (1200,2000), z \in (3, 5)} z/f\right)\bigg/\left(\min\limits_{f \in (1200,2000), z \in (3, 5)} z/f\right)} \approx 1.7
\end{align}

The data is fitted by the method by using the rendered image as input to a standard Resnet-50 with an output dimension of 7. Instead of applying a softmax on the output we use the mapping described in section \ref{subsec:parameter_remapping} to predict the parameters of \OurDist. Applying a negative log likelihood loss gives equation \ref{eq:loss_after_basis_change} which is used as the loss when fitting the parameters of the network. The parameters are updated using Adam with default parameters. This setup is visualized in figure \ref{fig:experimental_setup}

To showcase different cases we also generate 
additional datasets where the cylinder is scaled with a factor uniformly sampled from U(0.8,1.2). This dataset is constructed to showcase how the method performs when a scale/depth ambiguity is present.

We also generate a dataset where we also add a cube and a sphere to both add visual complexity and introduce occlusions which incentive the model to predict different uncertainties for the case where the cylinder is occluded compared to when it is not. For this dataset no scale ambiguity is present. The cube and sphere have orientations, material properties and color sampled indepenently from the cylinder and each other.

Finally we generate a dataset where we render a scene with occluding objects and with scale ambiguity. The scaling factor is sampled independently for all objects.

Each dataset containts 8000 training samples and 2000 test samples.
\subsection{Experimental result}

\subsubsection{Empirical error vs. estimated error}

We investigate how the predicted uncertainty correlates with the empirical error by for each sample computing the empirical squared error for the projected coordinates and for the depth. We then compute the predicted variance for each sample based on $A$ and $a$ by using equation \ref{eq:analytical_var_projected} and \ref{eq:analytical_var_depth}

We check if the estimated uncertainties are well correlated by sorting the samples based on the predicted variance and apply a low pass filter over 200 adjacent samples for both the predicted variance and the empirical squared error. For a perfect model these should be identical. For a good model they should be highly correlated.

\begin{figure}[h]
\includegraphics[width=\linewidth]{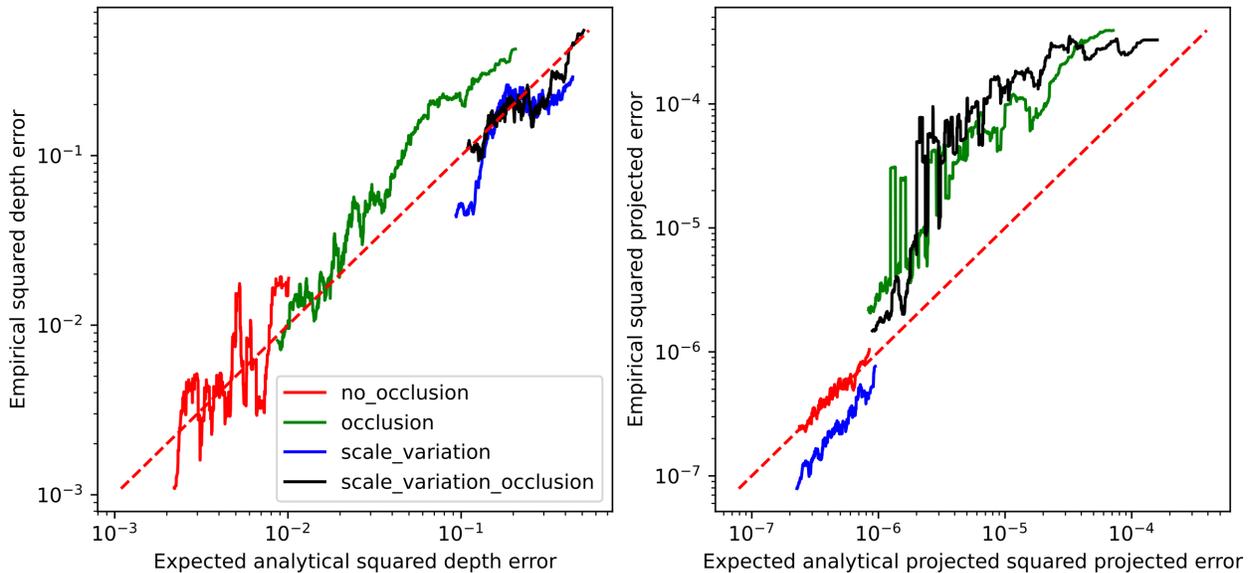}
\caption{correlation between predicted variance and empirical variance for depth (left) and projected coordinates (right). Red dashed line correspond to when the expected analytical depth and empirical squared error are equal. A method which predict uncertainty well should produce a curve which is close to this line. The unit for the left plot is meters, for the right plot the quantity is the unitless ratio between projected error and depth}
\label{fig:emp_vs_est_error}
\end{figure}
\begin{figure}[ht]
\includegraphics[width=\linewidth]{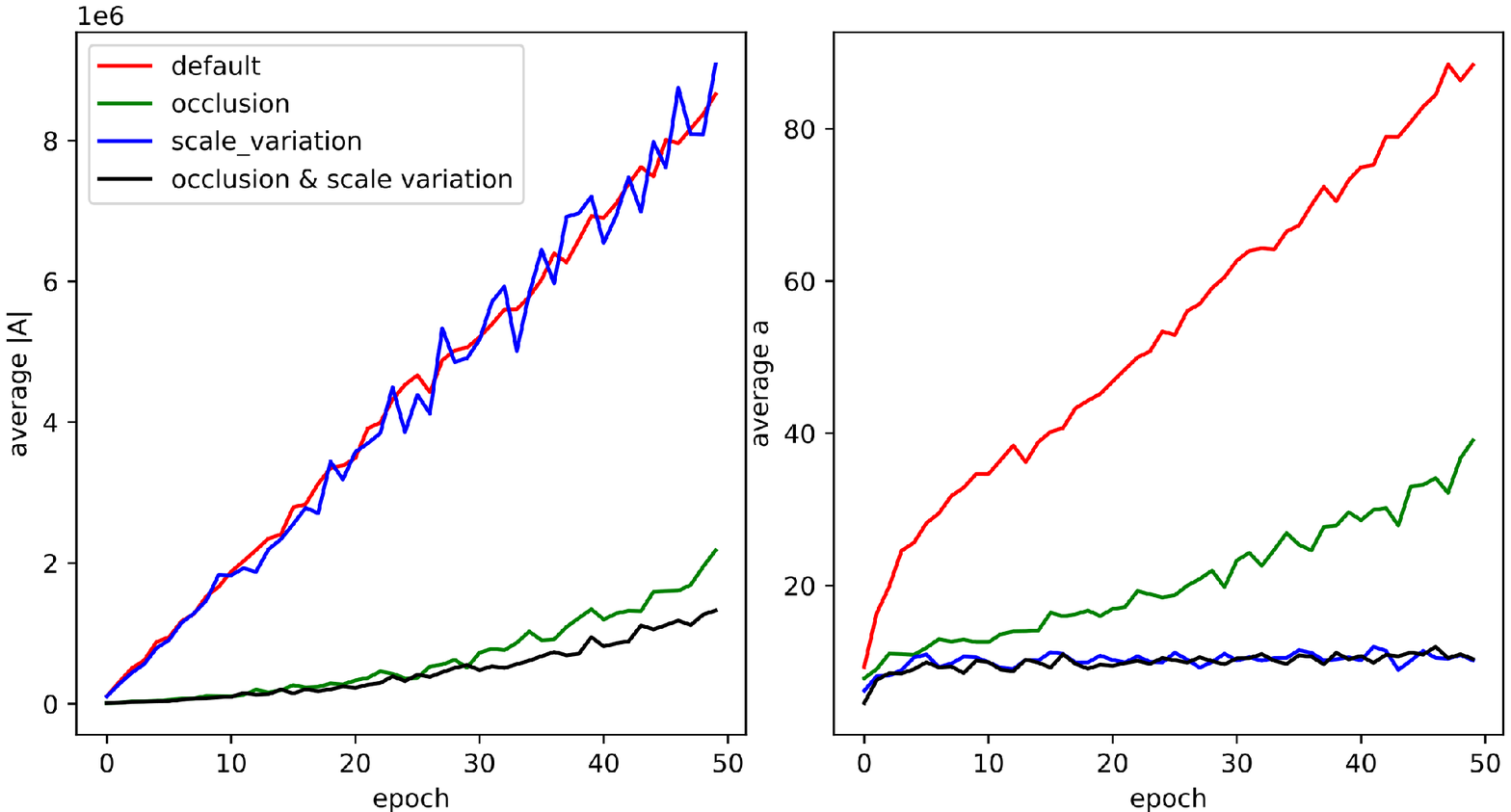}
\caption{Figure showing how the average precision parameters increase during the training. In the case when there are no occluding objects the estimated precision in image coordinates increase quickly during the training, with occlusion estimating the position in image space is harder and the precision increase slower. The precision of the depth estimate increases throughout training when there is no scale ambiguity, but converges to a constant when there is scale ambiguity }
\label{fig:evolution_of_a}
\end{figure}

From figure \ref{fig:emp_vs_est_error} we see that the estimated errors correlate well with the empirical errors over all datasets for both the depth and projected component. This indicates that the uncertainty estimates which the model produces are useful.
The estimated uncertainties also follow intuition, the dataset without occlusion or scale ambiguity has low predicted uncertainty for both the projected coordinates and for the depth.
For the dataset without occlusion but with scale ambiguity the depth uncertainty increases significantly, but the uncertainty for the projected coordinates remains low.
When fitting the model on a dataset without scale ambiguity but with other objects which can occlude the cylinder both the predicted uncertainty for the projected cordinates and the depth increases, which is expected since it is harder to predict where objects are if they are partially or fully occluded.

We also show how the average predicted parameter $a$ and $A$ change over training in figure \ref{fig:evolution_of_a}. From this figure we see that the predicted uncertainty decreases as the training proceeds. This is to due to the model adapt to fact that the error decrease when training.

\subsubsection{Regression performance}

To show that our method is able to estimate 3d position well we compare it against several regression baselines.
The reference is to try to regress $v_p$ and $z_p$ directly. Some methods estimate the logarithm of depth instead of the raw depth such as \cite{lin2020hdnet}. For completeness we compare to estimating $\log(z_p)$ as well.
The results of our method and these baselines is shown in table \ref{tab:ours_vs_baselines}. From this table we see that our method often exceeds the performance of both baselines, however this improvement is not likely to be significant, but does not need to be so either since our contribution is giving a probabilistic prediction with a convex negative log likelihood, not to produce a better single point estimate.
The errors are measured in the cameras coordinate system. For the baselines the predicted $(x,y,z)$ coordinates are computed by using the mapping in equation \ref{eqs:basis_change_variables}.
For our method we use the mode as our single point prediction
\begin{equation}
    (x,y,z) = \mu_z(\mu_x, \mu_y, 1)
\end{equation}
To construct a point estimate. There are other possible single point predictors which could be equally good, such as geometric median or mean we did not evaluate these, but they should produce similar predictions.

We also do ablations where we ignore the focal length of the camera when estimating the position, thereby violating our proposed constraint 2. One such baseline is to let the network predict $x,y,z$ directly using a MAE loss.
Another such baseline is to try to predict $v_p$ and $z_p$ but with a constant focal length of $f = \sqrt{f_{max}f_{min}} \approx 1550$ for the mapping between $x,y,z$ and $v_p, z_p$. Note that the camera used to capture the scene had different focal length for different samples, we only use an incorrect average focal length for the mapping in equation \ref{eqs:basis_change_variables} for training and evaluation. The results for the baselines and our method on the dataset without occluding objects nor scale ambiguity is shown in Table \ref{tab:normalized_vs_direct}.

From this table we see that baselines which ignore the focal length perform significantly worse, likely due to the fact that in general it is not possible to measure depth from images, only scale, therefore by ignoring the focal length an irreducible error is introduced.

Finally we try to frame how large our errors are in image space. On average the focal length is approximately 1550 pixels. The object is on average 4000 mm away. Therefore the average error of 2.4mm correspond to approximately $0.9$ pixels which quite good. Objects with a diameter of 0.2m will have a diameter of approximately 80 pixels when rendered with a focal length of 1550 at a depth of 4 meter. If estimating depth was done by the proxy task of estimating the diameter of objects an error of 1 pixel would therefore result in a depth error of 50mm which is on par with our average error when there are no occlusions or scale ambiguities, again indicating that our method works quite well.

Introducing a 20\% scale ambiguity of the rendered object should result in irreducible errors which are on average 10\% error of the depth. The object was on average 4000mm from the camera, which should give an average error of approximately 400mm. The measured errors which we observe are similar to this, but slightly lower, possibly due to the network being able to infer some information about the scale through some non-intuitive shading effect or due to the fact that $z_p$ does not follow a uniform distribution.

\begin{table}
\begin{tabular}{lc|cccr}
   Method & Metric & default & occl. & scale ambig.& occl. \& scale ambig \\
   \hline
   MAE, $v_p$, $z_p$ & projected error (mm)  & 3.6 & 22.8 & 3.7 & 37.2\\
   & depth error (mm)  & 69 & 218 & 359 & 386\\
   \hline
   MAE, $v_p$, $\log(z_p)$ & projected error (mm) & 5.0 & 18.0 & 5.1 & 27.6\\
   & depth error (mm)  & \textbf{36} & 178 & 355 & 355\\
   \hline
   probabilistic (ours) & projected error (mm)  & \textbf{2.4} & \textbf{13.0} & \textbf{1.6} & \textbf{17.0}\\
   & depth error (mm)  & 48 & \textbf{168} & \textbf{347} & \textbf{350}\\
\end{tabular}
\caption{regression performance of our proposed method (bottom) compared to two plain regression baselines}
\label{tab:ours_vs_baselines}
\end{table}

\begin{table}
\begin{tabular}{lcr}
   Method & Projected error & Depth error\\
   MAE, (x,y,z) & 13.0 & 327 \\
   MAE, $v_p$, $z_p$, constant $f$ & 10.8 & 325 \\
   probabilistic (ours) & \textbf{2.4} & \textbf{48} \\
\end{tabular}
\caption{Ablation on mean absolute error losses when 1) trying to directly regress $x,y,z$, 2) when regressing $v_p, z_p$ where the focal length is approximated with a constant and 3) When using our method}
\label{tab:normalized_vs_direct}
\end{table}

\section{Discussion}
\label{sec:discussion}
The constraints which we described are well motivated for the general case, but there are cases where they are not necessary. For example if scale ambiguity is not present in the dataset then constraint 1 is not necessary.
If the focal length is the same for all pictures in the dataset then constraint 2 is not neccessary.
It is also possible to infer camera intrinsics from some scenes if this is the case then constraint 2 might not be necessary either.
If the uncertainty in depth is small many distributions would assign a small probability that the object is behind the camera even if constraint 3 is not strictly true indicating that this constraint is mainly important when the depth uncertainty is large.
Constraint 5 is only necessary if the estimated probability distribution needs to be log concave.
we have shown this is useful for multi view fusion and imposing ground plane constraints, but if the goal is only to produce a as good as possible single point prediction of the position this constraint should not be necessary.

Furthermore our ablations indicate that using the correct focal length is important when estimating 3d position from camera data.

The focus of this work is to investigate the theoretical aspects of probabilistic 3d regression from camera data. To be able to highlight this the experiments are on synthetic data. Future work could be to verify that the proposed method works on more challenging real world data as well.

We also show that it should be easy to combine our method with multi-view fusion or the ground plane assumption in theory, but we do not implement it. This could also be future work.

\section{Conclusion}
\label{sec:conclusion}
In this work we have described constraints which we argue should be taken into account when designing a method to estimating probability distributions over 3d position from camera data. We have also designed a method which conform to these constraints. In our experiments we show that the uncertainty estimates which our method produce correlate well with empirical errors. Our experiments also show that our method perform on par or better than several regression baselines. In our ablations we show that in our experimental setting the performance decreases significantly if the camera intrinsics are ignored.

\section{Acknowledgements}
DM is an industrial PhD student at Tobii AB.
this work was partially supported by the Wallenberg AI, Autonomous Systems and Software Program
(WASP) funded by the Knut and Alice Wallenberg Foundation.

\bibliography{main}

\begin{thebibliography}{18}
\providecommand{\natexlab}[1]{#1}
\providecommand{\url}[1]{\texttt{#1}}
\expandafter\ifx\csname urlstyle\endcsname\relax
  \providecommand{\doi}[1]{doi: #1}\else
  \providecommand{\doi}{doi: \begingroup \urlstyle{rm}\Url}\fi

\bibitem[An(1997)]{an1997log}
Mark~Yuying An.
\newblock Log-concave probability distributions: Theory and statistical
  testing.
\newblock \emph{Duke University Dept of Economics Working Paper}, \penalty0
  (95-03), 1997.

\bibitem[Bertoni et~al.(2019)Bertoni, Kreiss, and Alahi]{Bertoni_2019_ICCV}
Lorenzo Bertoni, Sven Kreiss, and Alexandre Alahi.
\newblock Monoloco: Monocular 3d pedestrian localization and uncertainty
  estimation.
\newblock In \emph{Proceedings of the IEEE/CVF International Conference on
  Computer Vision (ICCV)}, October 2019.

\bibitem[Bruns \& Jensfelt(2022)Bruns and Jensfelt]{bruns2022evaluation}
Leonard Bruns and Patric Jensfelt.
\newblock On the evaluation of rgb-d-based categorical pose and shape
  estimation.
\newblock \emph{arXiv preprint arXiv:2202.10346}, 2022.

\bibitem[Choi et~al.(2016)Choi, Zhou, Miller, and Koltun]{choi2016large}
Sungjoon Choi, Qian-Yi Zhou, Stephen Miller, and Vladlen Koltun.
\newblock A large dataset of object scans.
\newblock \emph{arXiv preprint arXiv:1602.02481}, 2016.

\bibitem[Feng et~al.(2019)Feng, Rosenbaum, Glaeser, Timm, and
  Dietmayer]{feng2019can}
Di~Feng, Lars Rosenbaum, Claudius Glaeser, Fabian Timm, and Klaus Dietmayer.
\newblock Can we trust you? on calibration of a probabilistic object detector
  for autonomous driving.
\newblock \emph{arXiv preprint arXiv:1909.12358}, 2019.

\bibitem[Geiger et~al.(2012)Geiger, Lenz, and Urtasun]{Geiger2012CVPR}
Andreas Geiger, Philip Lenz, and Raquel Urtasun.
\newblock Are we ready for autonomous driving? the kitti vision benchmark
  suite.
\newblock In \emph{Conference on Computer Vision and Pattern Recognition
  (CVPR)}, 2012.

\bibitem[Ionescu et~al.(2013)Ionescu, Papava, Olaru, and
  Sminchisescu]{ionescu2013human3}
Catalin Ionescu, Dragos Papava, Vlad Olaru, and Cristian Sminchisescu.
\newblock Human3. 6m: Large scale datasets and predictive methods for 3d human
  sensing in natural environments.
\newblock \emph{IEEE transactions on pattern analysis and machine
  intelligence}, 36\penalty0 (7):\penalty0 1325--1339, 2013.

\bibitem[Iskakov et~al.(2019)Iskakov, Burkov, Lempitsky, and
  Malkov]{iskakov2019learnable}
Karim Iskakov, Egor Burkov, Victor Lempitsky, and Yury Malkov.
\newblock Learnable triangulation of human pose.
\newblock In \emph{Proceedings of the IEEE/CVF International Conference on
  Computer Vision}, pp.\  7718--7727, 2019.

\bibitem[Johnson et~al.(2017)Johnson, Hariharan, Van Der~Maaten, Fei-Fei,
  Lawrence~Zitnick, and Girshick]{johnson2017clevr}
Justin Johnson, Bharath Hariharan, Laurens Van Der~Maaten, Li~Fei-Fei,
  C~Lawrence~Zitnick, and Ross Girshick.
\newblock Clevr: A diagnostic dataset for compositional language and elementary
  visual reasoning.
\newblock In \emph{Proceedings of the IEEE conference on computer vision and
  pattern recognition}, pp.\  2901--2910, 2017.

\bibitem[Kumar et~al.(2020)Kumar, Marks, Mou, Wang, Jones, Cherian,
  Koike-Akino, Liu, and Feng]{kumar2020luvli}
Abhinav Kumar, Tim~K Marks, Wenxuan Mou, Ye~Wang, Michael Jones, Anoop Cherian,
  Toshiaki Koike-Akino, Xiaoming Liu, and Chen Feng.
\newblock Luvli face alignment: Estimating landmarks' location, uncertainty,
  and visibility likelihood.
\newblock In \emph{Proceedings of the IEEE/CVF Conference on Computer Vision
  and Pattern Recognition}, pp.\  8236--8246, 2020.

\bibitem[Lin \& Lee(2020)Lin and Lee]{lin2020hdnet}
Jiahao Lin and Gim~Hee Lee.
\newblock Hdnet: Human depth estimation for multi-person camera-space
  localization.
\newblock In \emph{European Conference on Computer Vision}, pp.\  633--648.
  Springer, 2020.

\bibitem[Meyer et~al.(2019)Meyer, Laddha, Kee, Vallespi-Gonzalez, and
  Wellington]{meyer2019lasernet}
Gregory~P Meyer, Ankit Laddha, Eric Kee, Carlos Vallespi-Gonzalez, and Carl~K
  Wellington.
\newblock Lasernet: An efficient probabilistic 3d object detector for
  autonomous driving.
\newblock In \emph{Proceedings of the IEEE/CVF conference on computer vision
  and pattern recognition}, pp.\  12677--12686, 2019.

\bibitem[Mills \& Goldenberg(1989)Mills and Goldenberg]{mills1989force}
James~K Mills and Andrew~A Goldenberg.
\newblock Force and position control of manipulators during constrained motion
  tasks.
\newblock \emph{IEEE Transactions on Robotics and Automation}, 5\penalty0
  (1):\penalty0 30--46, 1989.

\bibitem[Mohlin et~al.(2021)Mohlin, Bianchi, and
  Sullivan]{mohlin2021probabilistic}
David Mohlin, Gerald Bianchi, and Josephine Sullivan.
\newblock Probabilistic regression with huber distributions.
\newblock \emph{arXiv preprint arXiv:2111.10296}, 2021.

\bibitem[Moon et~al.(2019)Moon, Chang, and Lee]{moon2019camera}
Gyeongsik Moon, Ju~Yong Chang, and Kyoung~Mu Lee.
\newblock Camera distance-aware top-down approach for 3d multi-person pose
  estimation from a single rgb image.
\newblock In \emph{Proceedings of the IEEE/CVF international conference on
  computer vision}, pp.\  10133--10142, 2019.

\bibitem[von Marcard et~al.(2018)von Marcard, Henschel, Black, Rosenhahn, and
  Pons-Moll]{vonMarcard2018}
Timo von Marcard, Roberto Henschel, Michael Black, Bodo Rosenhahn, and Gerard
  Pons-Moll.
\newblock Recovering accurate 3d human pose in the wild using imus and a moving
  camera.
\newblock In \emph{European Conference on Computer Vision (ECCV)}, sep 2018.

\bibitem[Wang et~al.(2022)Wang, Xinge, Pang, and Lin]{wang2022probabilistic}
Tai Wang, ZHU Xinge, Jiangmiao Pang, and Dahua Lin.
\newblock Probabilistic and geometric depth: Detecting objects in perspective.
\newblock In \emph{Conference on Robot Learning}, pp.\  1475--1485. PMLR, 2022.

\bibitem[Zhen et~al.(2020)Zhen, Fang, Sun, Liu, Jiang, Bao, and
  Zhou]{zhen2020smap}
Jianan Zhen, Qi~Fang, Jiaming Sun, Wentao Liu, Wei Jiang, Hujun Bao, and
  Xiaowei Zhou.
\newblock Smap: Single-shot multi-person absolute 3d pose estimation.
\newblock In \emph{European Conference on Computer Vision}, pp.\  550--566.
  Springer, 2020.

\end{thebibliography}
\bibliographystyle{tmlr}

\newpage
\appendix
\section{Appendix}
\section{Projected and depth independent result in finite precision}
\label{sec:supp_limited_precision_for_decompose_vpz}
\subsection{Variance decreases when reducing size of single tail}
This section proves an intermediate result which we use for the main proof
\label{sec:supp_c_decreases_variance}
Here we show that for random varianbles Y,Z and a variable $c>0$ and $p \in [0,1]$
then if $E\left[Y\right]E\left[Z\right] < 0$ then 
\begin{equation}
    X = \begin{cases}
    cY \text{ w.p } p \\
    Z \text{ otherwise }
    \end{cases}
\end{equation}
Then $Var(X)$ decreases in $c$ decreases

\textbf{proof}

\begin{align}
    E\left[X\right] &= pcE\left[Y\right] + (1-p)E\left[Z\right]\\
    E\left[X^2\right] &= pc^2E\left[Y^2\right] + (1-p)E\left[Z^2\right]\\
    Var\left[X\right] &= c^2p(E\left[Y^2\right]-(E\left[Y\right])^2) + c^2(p-p^2)(E\left[Y\right])^2 - 2cp(1-p)E\left[Y\right]E\left[Z\right] + ...
\end{align}

Where ... does not depend on c. The first term is $c^2$ time the variance of Y which is positive, the second term is positive since $(p-p^2) > 0$. The third term is positive since $E\left[Y\right]E\left[Z\right] < 0$ and the negations cancel out.

Therefore if $c > 0$ $Var(X)$ increase when $c$ increase, which conclude the proof.

\subsection{Main proof}

Proof sketch, assume all constraints except constraint 1 are true, show that the smallest possible variance for the projected coordinates is at least a constant which is larger than 0.

By considering the line $y=0$ we get
\begin{equation}
    P(x,y=0,z) = p_{proj}(x/z,0)p_{depth}(z)
\end{equation}

The negative log likelihood is

\begin{equation}
    -\log(p(x,y=0,z)) = -\log(p_{proj}(x/z,0))-\log(p_{depth}(z))
\end{equation}

For convenience we define 
\begin{align}
    f(x/z) &= -\log(p_{proj}(x/z,0)) \\
    g(z) &= -\log(p_{depth}(z)
\end{align}

We compute the hessian with respect to $x$ and $z$ of

\begin{align}
    \dfrac {\partial } {\partial x} -\log(p(x,y=0,z)) &= f'(x/z)/z\\
    \dfrac {\partial^2 } {\partial x^2} -\log(p(x,y=0,z)) &= f''(x/z)/z^2\\
    \dfrac {\partial^2 } {\partial x\partial z} -\log(p(x,y=0,z)) &= -f'(x/z)/z^2-f''(x/z)x/z^3\\
    \dfrac {\partial} {\partial z} -\log(p(x,y=0,z)) &= g'(z) -f'(x/z)x/z^2\\
    \dfrac {\partial^2} {\partial z^2} -\log(p(x,y=0,z)) &= g''(z) +f''(x/z)x^2/z^4+2f'(x/z)x/z^3
\end{align}

If this function is convex then this hessian has a positive determinant.
The determinant is

\begin{align}
    |H_{x,z}| &= \dfrac {1}{z^4} (f''(x/z) g''(z) + (f''(x/z))^2(x/z)^2+2f'(x/z)f''(x/z)x/z-(f'(x/z)+f''(x/z)(x/z))^2)\\
    &= \dfrac{1}{z^4}(f''(x_p)g''(z) - (f'(x_p))^2
\end{align}

The function is twice differentiable at least for one point, let $C = g''(z)$ exist for the $z$ value at this point. If $C<0$ Then we reach a contradition, if $C=0$ then $f'(x_p)=0$ Where the function has support. The solution here would be $p_{proj}(x_p) = 1/S for |x_p| \le S/2$ which can not model arbitrary precision.

The solution of

\begin{equation}
    Cf''(x_p) - f'(x_p)^2 = 0
\end{equation}
is 
\begin{equation}
    f(x_p) = -a \log(c_1 -c_2x)
\end{equation}

If the constraint is not tight, then the resulting function would increase faster than the solution of the differential equation as the distance to the mode increases.
If we study the function for $x_p$ which are larger than the mode then the derivative is
\begin{equation}
    f'(x_p) = a \dfrac {c_2} {c_1 -c_2x} 
\end{equation}
Since we are to the right of the mode $f'(x_p) > 0$, $c_1-c_2x$ is also positive since the logarithm of this value is real. Thus in this region $c_2 > 0$ Therefore $f$ approach infinity when $x \rightarrow c_1/c_2$. Therefore $c_1/c_2 \ge S/2$.

The mode has to exist since the distribution is proper.
The same analysis holds to show that for $x_p < m$

\begin{equation}
    f(x_p) = -C \log(b_1 +b_2x)
\end{equation}
Where $b_2 > 0$ and $b_1/b_2 > S/2$.

The distribution will either give a value less than the mode or larger or equal to the mode
We can now apply the proof in \ref{sec:supp_c_decreases_variance} to show $c_1/c_2 = S/2$ and $b_1/b_2 = S/2$
because if that was not the case it would be possible to reduce the variance by making the constraint active.

Furthermore $f$ is continuous since it is convex.

Thus the distribution which minimizes the projected variance while being consistent with the constraints has the form
\begin{equation}
    p_{proj}(x_p) \propto
    \begin{cases}
    (S/2-x)^C \text{ if } x > m \\
    \left(\dfrac {(S/2+x)(S/2-m)} {S/2+m}\right)^C \text{ otherwise}
    \end{cases}
\end{equation}

Which obviously can not model an arbitrary small variance for the projected coordinates for a fixed C.

Finally we need to show that 
\begin{equation}
    g''(z) \ge C
\end{equation}

Imposes constraints on what variance for z can be modeled

To maximize variance we minimize g while by keeping the constraint tight. This gives
\begin{equation}
g(z) = Cz^2/2+k_1z+k_2
\end{equation}

which gives the distribution after reparameterization

\begin{equation}
    p_{depth}(z) \propto \exp(-\dfrac {(z-\mu)^2} {2/C})
\end{equation}
Which is a normal distribution with variance $1/\sqrt{C}$

Therefore a low variance in the projected coordinates requires a large $C$ which requires a low variance for the depth estimate. This concludes the proof.

\section{Multi-view fusion and ground plane constraints}
\label{sec:supp_multi_view_ground_plane}
In this section we prove proposition 2 and 3.

\subsection{Multi view}

For this section we have $n$ cameras which produce predictions
$p_{cam}(v_i; \theta_i)$ where $v_i$ are coordinates in the coordinate system for camera $i$. $\theta_i$ are the parameters of the estimated probability distribution based on the image captured by camera $i$. The images are in theory captured at the same time or in practice within a very short time period.
The multi view fusion is constructed under the assumption that the errors for each estimate are independent. The affine transform $v_1 = R_iv_i+t_i$ transform coordinates from coordinate system $i$ to coordinate system $1$

\textbf{Definition 1:} We define the maximum likelihood fusion given the estimates to be
\begin{equation}
   \arg\sup\limits_{v_1 \in \mathbb{R}^3} p(v_1|\theta_1, \cdots \theta_n) = \prod\limits_{i=1}^n p(R_i^{-1}(v_1-t_i)|\theta_i)
\end{equation}

We introduce the notion of a valid fusion as

\textbf{Definition 2:} A valid maximum likelihood fusion point exist if
\begin{equation}
   (\exists v\in \mathbb{R}^3 \text{ s.t. } p_{cam}(v | \theta_1, \cdots \theta_n) = \prod\limits_{i=1}^n p_{cam}(R_i^{-1}(v_1-t_i)|\theta_i))\land (\forall i \in \{1, \cdots n\} p_{cam}(R_i^{-1}(v_1-t_i)|\theta_i) > 0)
\end{equation}

To exclude the case where either most likely fusioned point exist infinitely far away or when the fusioned distribution is improper.

\textbf{Proposition: 2 (existance)} In multi view fusion, if the field of view for the cameras have a non-empty intersection then if each camera gives probability estimates conforming to constraints 1-7 there will exist a valid maximum likelihood fusion point.

\textbf{Proof}
Having a non empty intersection of the field of views for the cameras implies
\begin{equation}
\exists v_1 \text{ such that } \forall i \in \{1, \cdots n\}
v_i = R_i^{-1}(v_1-t_i)
\text{ and } v_i \in FOV_{camera,i}
\end{equation}

Constraint 4 implies that $p_{cam}(v_i|\theta_i) = p_{cam}(R_i^{-1}(v_1-t_i)|\theta_i) > 0$ for all cameras. since $v_i$ is in the field of view of the camera

Therefore 
\begin{equation}
p(v_1|\theta_1, \cdots \theta_n) = \prod\limits_{i=1}^n p_{cam}(R_i^{-1}(v_1-t_i)|\theta_i) > 0
\end{equation}

Constraint 6 implies that there exist a $R$ such that
\begin{equation}
\forall v \text{ such that if } \|v\| > R \text{ then } p(v|\theta_1 \cdots \theta_n) < p(v_1|\theta_1, \cdots \theta_n)
\end{equation}

therefore 
\begin{align}
\sup\limits_{v} p(v|\theta_1, \cdots \theta_n)
&= \sup\limits_{v \text{ s.t. } \|v\|\le R} p(v|\theta_1, \cdots \theta_n) \\
&= \max\limits_{v \text{ s.t. } \|v\|\le R} p(v|\theta_1, \cdots \theta_n)
\end{align}
The last step is due to
1) $(v|\theta_1, \cdots \theta_n)$ is continuous since continuity is preserved when multiplying continuous functions.\\
2) A closed ball is compact. \\
3) the extreme value theorem states that the maximum over a compact set of a continous function exists. Thus applying the extreme value theorem concludes the proof.

\textbf{Proposition 2: (convexity)} Finding the maximum likelihood fusion is a convex optimization problem.

\textbf{proof}
Maximizing a probability is equivalent to minimizing a log likelihood.

Formally, given definition 1

\begin{equation}
   p(v_1|\theta_1, \cdots, \theta_n) = \prod\limits_{i=1}^n p(R_i^{-1}(v_1-t_i)|\theta(im_i))
\end{equation}
Here we have used the fact that the transform from $v_1$ to $v_i$ has a jacobian with determinant 1.

Maximizing the probability is the same as minimizing a negative log likelihood, since logarithms are increasing and negating an increasing function result in a decreasing function, that is 
\begin{align}
\argmax\limits_{v_1 \in \mathbb{R}^3} \prod\limits_{i=1}^n p(R_i^{-1}(v_1-t_i)|\theta_i)
&= \argmin\limits_{v_1 \in \mathbb{R}^3} -\log(\prod\limits_{i=1}^n p_{cam}(R_i^{-1}(v_1-t_i)|\theta_i)) \\
&= \argmin\limits_{v_1 \in \mathbb{R}^3}\sum\limits_{i=1}^n -\log(p_{cam}(R_i^{-1}(v_1-t_i)|\theta_i))
\end{align}

It is also well known that convexity is preserved under linear transformations of the argument

This means that if $-\log (p_{cam}(v_i|\theta_i))$ is convex with respect to $v_i$ then $-\log (p(R_i^{-1}(v_1-t_i)|\theta_i))$ is convex with respect to $v_1$ since the change in coordinate system is a linear transform.

Convexity is also preserved under addition therefore if $-\log(p(R_i^{-1}(v_1 - t_i)|\theta_i))$ is convex with respect to $v_1$ for all $i$, then 
$\sum\limits_{i=0}^n -\log(p(R_i^{-1}(v_1 - t_i)|\theta_i))$ is convex with respect to $v_1$.

Finding a minima with respect to a convex function is a convex optimization problem. This concludes the proof.

\textbf{proposition 3:}
The constraints makes finding the most likely point while enforcing a ground plane assumption, that is solving 
\begin{equation}
\argmax_{v \text{ s.t. } d^Tv = c} p(v|\theta)
\end{equation}
where $d$ and $c$ define the ground plane is a convex optimization problem.
If any point of the ground plane is in the field of view of the camera, then the constraints also guarantee that a solution exists.

\textbf{proof}
\begin{align}
\argmax_{v \text{ s.t. } d^Tv = c} p(v|\theta)
&= \argmin_{v \text{ s.t. } d^Tv = c} -\log(p(v|\theta))
\end{align}

optimizing a convex function with an affine constraint in a convex optimization problem.

If a point $v_1$ of the ground plane is in the field of view of the camera, then constraint 4 ensures $p(v_1|\theta) > 0$

Constraint 6 ensures
\begin{equation}
\exists R \text{ s.t. } \max\limits_{v \text {s.t.} \|v\| > R} p(v|\theta) < p(v_1|\theta) 
\end{equation}

Therefore 
\begin{equation}
\argmax\limits_v p(v|\theta) = \argmax\limits_{v \text{ s.t. } \|v\| \le R} p(v|\theta)\end{equation}

Which exist due to extreme value theorem and constraint 7.

\section{Intermediate integrals}
\subsection{Double exponential integral}
For convenience for future proofs we derive the integral of a double exponential
\begin{equation}
\label{eq:double_exp_gamma}
\int\limits_{0}^{\infty} \exp(ax-c\exp(x)) dx = \Gamma(a, c) c^{-a}
\end{equation}
Where $\Gamma$ is the upper incomplete gamma function

\textbf{Proof}
\begin{align}
\int\limits_{0}^{\infty} \exp(ax-c\exp(x)) dx 
&=
\int\limits_{0}^{\infty} \exp(-alog(c)) \exp(a(x+log(c)))\exp(-\exp(x+log(c)))dx \\
&=
c^{-a} \int\limits_{log(c)}^{\infty}  \exp(ay)\exp(-\exp(y))dy \\
&=
c^{-a} \int\limits_{c}^{\infty} z^{a-1} \exp(-z)dz \\
&= c^{-a}\Gamma(a, c)
\end{align}

The first step is rearranging.

The second step comes from the variable substitution

\begin{equation}
    y = x+\log(c)
\end{equation}

The third step comes from the variable substitution

\begin{equation}
    z = \exp(y)
\end{equation}

\subsection{Moments depth distribuion}
Here we show
\begin{equation}
    \label{eq:moments_of_double_exp}
    \int\limits_{0}^{\infty} z^k \exp(-a\max(\mu/z, z/\mu))dz = \mu^{k+1}(\Gamma(-k-1, a)a^{k+1} + \Gamma(k+1, a)a^{-k-1})
\end{equation}

\textbf{proof}

\begin{align}
    \int\limits_{0}^{\infty} z^k \exp(-a\max(\mu/z, z/\mu))dz &= \int\limits_{0}^{\infty} z^{k+1} \dfrac {1}{z} \exp(-a\exp(|\log(z/\mu)|)) dz \\
    &= \mu^{k+1}\int\limits_{0}^{\infty} \left(\dfrac {z}{\mu}\right)^{k+1} \dfrac {1}{z} \exp(-a\exp(|\log(z/\mu)|)) dz \\
    &=
    \mu^{k+1}\int\limits_{-\infty}^{\infty}\exp((k+1)s)\exp(-a\exp(|s|))ds \\
    &=
    \mu^{k+1}(\int\limits_{0}^{\infty}\exp(-(k+1)s)\exp(-a\exp(s))ds\\
    &+\int\limits_{0}^{\infty}\exp((k+1)s)\exp(-a\exp(s))ds) \\
    &= \mu^{k+1}(\Gamma(-k-1, a)a^{k+1} + \Gamma(k+1, a)a^{-k-1})
\end{align}

The first step is rearranging. The second step is the variable substitution $s = \log(z/\mu)$. The third step is separating the range of the integral into positive and negative numbers. The final step is using equation \ref{eq:double_exp_gamma}.

\subsection{Expected value of a positive huber distribution in 1 dimension}
For use in future parts we derive the integral of
\begin{equation}
    \label{eq:expected_positive_huber}
    \int\limits_{0}^{\infty} r \exp(-h(r)) dr = 1 + \exp(-1/2)
\end{equation}

\textbf{Proof}
\begin{align}
    \int\limits_{0}^{1} r \exp(-r^2/2)) dr &= 
    \int\limits_{0}^{1/2} \exp(-y) dy \\
    &= 1-\exp(-1/2)
\end{align}
The first step is the variable substitution $y = r^2/2$, which has a scale factor of $dy/dr = r$

\begin{align}
    \int\limits_{1}^{\infty} r \exp(-r+1/2) dr &= 
    \left[-r\exp(-r+1/2)\right]_{1}^{\infty} + \int\limits_{1}^{\infty} \exp(-r+1/2) dr \\
    &= 2\exp(-1/2)
\end{align}
The first step is integration by parts.

This gives

\begin{align}
    \int\limits_{0}^{\infty} r \exp(-h(r)) dr &= 
    \int\limits_{0}^{1} r \exp(-r^2/2) dr + \int\limits_{1}^{\infty} r \exp(-r+1/2) dr \\
    &= 1+\exp(-1/2)
\end{align}

\subsection{Expected cube of positive Huber distribution in 1 dimension}
For use in future parts we derive the integral of
\begin{equation}
    \label{eq:cube_positive_huber}
    \int\limits_{0}^{\infty} r^3 \exp(-h(r)) dr = 4+3\exp(-1/2)
\end{equation}

\textbf{Proof}
\begin{align}
\int\limits_{0}^{\infty} r^3 \exp(-h(r)) dr &= \int\limits_{0}^{1} r^3 \exp(-r^2/2) dr + 
\int\limits_{1}^{\infty} r^3 \exp(-r+1/2) dr
\end{align}

\begin{align}
\int\limits_{0}^{1} r^3 \exp(-r^2/2) dr &= \int\limits_{0}^{1} r^2 r \exp(-r^2/2) dr \\
&= \int\limits_{0}^{1/2} 2y \exp(-y) dy \\
&= [-2y\exp(-y)]_0^{1/2} +\int\limits_{0}^{1/2} 2\exp(-y) dy \\
&= 2 - \exp(-1/2) + 2 - 2\exp(-1/2)\\
&= 4-3\exp(-1/2)
\end{align}

\begin{align}
\int\limits_{1}^{\infty} r^3 \exp(-r+1/2) dr 
&= \exp(1/2)([-r^3\exp(-r)]_1^{\infty}
+\int\limits_{1}^{\infty} 3r^2\exp(-r) dy) \\
&= \exp(1/2)([-3r^2\exp(-r)]_1^{\infty}
+\int\limits_{1}^{\infty} 6r\exp(-r) dy) \\
&= \exp(1/2)([-6r^2\exp(-r)]_1^{\infty}
+\int\limits_{1}^{\infty} 6\exp(-r) dy) \\
&= 6\exp(-1/2)
\end{align}

Therefore 
\begin{equation}
    \int\limits_{0}^{\infty} r^3 \exp(-h(r)) dr = 4+3\exp(-1/2)
\end{equation}

\subsection{Norm factor of 2d Huber distribution}
\begin{equation}
K_{huber} = 2\pi(1+\exp(-1/2))
\end{equation}

\textbf{ proof}

\begin{align}
K_{huber} = \int\limits_{\mathbb{R}^2} \exp(-h(\sqrt{x^2 + y^2})) dxdy = 2\pi\int\limits_{0}^{\infty} r\exp(-h(r))dr = 2\pi(1+\exp(-1/2))
\end{align}

\subsection{Covariance of 2d Huber distribution}

\begin{equation}
    E[vv^T] = I\dfrac{(4+3\exp(-1/2))} {2+2\exp(-1/2)}
\end{equation}

\textbf{proof}

\begin{align}
    E[tr(vv^T)] = \dfrac{1} {K_{huber}}\int_{\mathbb{R}^2} v^Tv \exp(-h(\|v\|_2)) dv = \\
    \dfrac{1} {K_{huber}}\int_0^{2\pi}\int_{0}^{\infty} r^3 \exp(-h(r)) dr = \dfrac{4+3\exp(-1/2)} {1+\exp(-1/2)}
\end{align}

From rotation symmetry we know the variance will be a scaled identity matrix

Therefore 
\begin{equation}
    E[vv^T] = I \dfrac {4+3\exp(-1/2)} {2+2\exp(-1/2)}
\end{equation}

\section{\OurDist components}
\label{sec:supp_components_moments}
\subsection{Depth distribution}

\begin{equation}
p_{depth}(z;a,\mu_z) = \begin{cases}\dfrac {1} {K_{depth}(a,\mu_z)}\exp(-a \max(z/\mu_z, \mu_z/z)) &\text {if z < 0} \\
0 &\text{otherwise}
\end{cases}
\end{equation}

with 
\begin{equation}
K_{depth}(a, \mu_z) = \mu_z(exp(-a)/a + \Gamma(-1, a)a)
\end{equation}

\textbf{proof}
The normalizing factor can be computed as 
\begin{align}
   K_{depth}(\mu,a) &= \int\limits_{0}^{\infty}\exp(-a \max(z/\mu_z, \mu_z/z))dz \\
   &=
   \mu_z(\Gamma(1, a)/a + \Gamma(-1, a)a) \\
   &= \mu_z(exp(-a)/a + \Gamma(-1, a)a)
\end{align}

The first step is from definition.
The second step is using equation \ref{eq:moments_of_double_exp}. The third step is 

\begin{equation}
    \Gamma(1,a) = \int\limits_{a}^{\infty} t^{1-1}\exp(-t) dt = \exp(-a)
\end{equation}

\subsubsection{Expected value of depth}
\begin{equation}
    E\left[z\right] = \mu_z \dfrac {\Gamma(2, a)a^{-2} + \Gamma(-2, a)a^2}{\exp(-a)a^{-1} + \Gamma(-1, a)a}
\end{equation}
This function is decreasing with respect to a
with limit $\mu_z$ as $a \rightarrow \infty$
When $a>1$, $\mu_z < E\left[z\right] < 1.7\mu_z$

\textbf{proof}

Follows directly from $K_{depth}$ and equation \ref{eq:moments_of_double_exp}

The limit can be proven by multiplying numerator and denominator by $a\exp(a)$
This gives

\begin{align}
    \lim\limits_{a \rightarrow \infty} E[z;a,\mu_z] = \lim\limits_{a \rightarrow \infty} \mu_z \dfrac {\Gamma(2,a)\exp(a)/a+\Gamma(-2,a)\exp(a)a^3} {1+\Gamma(-1,a)\exp(a)a^2}
    &= \mu_z \dfrac {1+1} {1+1} = \mu_z
\end{align}

Since
\begin{equation}
\lim\limits_{a \rightarrow \infty} \Gamma(k,x)x^{-k+1}\exp(x) = 1
\end{equation}
\subsubsection{Variance of depth}
The variance of the depth is 

\begin{equation}
    Var \left[z\right] = \mu_z^2 \dfrac {(\Gamma(1,a)/a+\Gamma(-1,a)a)(\Gamma(3,a)a^{-3}+\Gamma(-3,a)a^{3}) - (\Gamma(2,a)a^2+\Gamma(-2,a)a^{-2})^2} {(\Gamma(1, a)/a + \Gamma(-1, a)a)^2}
\end{equation}

This expression is not very intuitive, but it behaves like
\begin{equation}
    Var\left[z\right] \approx \mu_z/a^2
\end{equation}

when $a > 1$

\textbf{proof}

\begin{equation}
 E\left[z^2\right] = \mu_z^2\dfrac {\Gamma(3,a)a^{-3}+\Gamma(-3,a)a^{3}} {\Gamma(1, a)/a + \Gamma(-1, a)a}
\end{equation}
Follows directly from $K_{depth}$ and equation \ref{eq:moments_of_double_exp}

\begin{equation}
 Var\left[z^2\right] = E\left[z^2\right] - E\left[z\right]^2
\end{equation}

\subsection{Projected distribution}

The normalizing factor of the projected part of the distribution is 

\begin{equation}
    K_{proj}(A, \mu) = \dfrac {\mu_z^2} {|A|} 2\pi(1+\exp(-1/2))
\end{equation}

For a distribution over a given plane where z is constant.

\textbf{Proof}

\begin{align}
\int_{-\infty}^{\infty}\int_{-\infty}^{\infty} \exp\left(-h\left(\left\|A\begin{bmatrix}
x/z-\mu_x\\
y/z-\mu_y
\end{bmatrix}\right\|_2\dfrac {z} {\mu_z}\right)\right) dxdy
&= \int_{-\infty}^{\infty}\int_{-\infty}^{\infty} \exp\left(-h\left(\left\|A\begin{bmatrix}
x/\mu_z-\mu_xz/\mu_z\\
y/\mu_z-\mu_yz/\mu_z
\end{bmatrix}\right\|_2\right)\right) dxdy \\
&= \dfrac {\mu_z^2} {|A|} \int_{-\infty}^{\infty}\exp(-h(\sqrt{x_p^2+y_p^2}))dx_pdy_p \\
&= 2\pi (1+\exp(-1/2))\dfrac {\mu_z^2} {|A|}
\end{align} 
The second step is the variable change
\begin{equation}
    \begin{bmatrix} x_p \\ y_p\end{bmatrix}
    = A\begin{bmatrix} x/\mu_z - \mu_xz/\mu_z \\ y/\mu_z - \mu_yz/\mu_z
    \end{bmatrix} 
\end{equation}
The last step comes from \ref{eq:expected_positive_huber} after a variable change to polar coordinates.

\subsection{Expected value of projected components}

The expected value

\begin{equation}
E_{x,y}\begin{bmatrix} x/z \\ y/z \end{bmatrix}
= \begin{bmatrix} \mu_x \\ \mu_y \end{bmatrix}
\end{equation}

for a fixed plane where $z$ is constant.

\textbf{Proof}
\begin{align}
E_{x,y}\begin{bmatrix} x/z \\ y/z \end{bmatrix} &= \dfrac {1} {K_{proj}(A, \mu)} \int_{-\infty}^{\infty}\int_{-\infty}^{\infty} \begin{bmatrix} x/z \\ y/z \end{bmatrix}  \exp\left(-h\left(\left\|A\begin{bmatrix}
x/z-\mu_x\\
y/z-\mu_y
\end{bmatrix}\right\|_2\dfrac {z} {\mu_z}\right)\right) dxdy \\ 
&= \dfrac {|A|} {K_{proj}(A, \mu)\mu_z^2} \int_{-\infty}^{\infty}\int_{-\infty}^{\infty} (\dfrac{\mu_z} {z} A^{-1} v_p +\mu_p) \exp\left(-h\left(\|v_p\|_2\right)\right) dx_pdy_p \\
&= \mu_p
\end{align}

where the basis change is the same as when deriving the normalizing factor. The last step is recognizing that the expected value of $v_p = \bar{0}$ due to symmetry.

\subsection{variance of projected components}
The for a constant z value is
\begin{equation}
Var_{x,y}\begin{bmatrix} x/z \\ y/z \end{bmatrix} = \dfrac {\mu_z^2} {z^2} A^{-2} \dfrac {4+3\exp(-1/2)}{2+2\exp(-1/2)}
\end{equation}

\textbf{proof}
\begin{align}
E_{x,y}\left[\begin{bmatrix} x/z \\ y/z \end{bmatrix}\begin{bmatrix} x/z \\ y/z \end{bmatrix}^T\right] &= E_{x,y}\left[ (\dfrac {\mu_z} {z} A^{-1}v_p + \mu_p)(\dfrac {\mu_z} {z} A^{-1}v_p + \mu_p)^T\right] \\
&= (\dfrac {\mu_z^2} {z^2}A^{-1}E\left[v_pv_p^T\right]A^{-1} + 
\dfrac {\mu_z} {z}A^{-1}E\left[v_p\right]\mu_p^T + 
\dfrac {\mu_z} {z}\mu_pE\left[v_p^T\right]A^{-1} + \mu_p\mu_p^T \\
&= \dfrac {\mu_z^2} {z^2} A^{-1}A^{-1} \dfrac {4+3\exp(-1/2)}{2+2\exp(-1/2)} + \mu_p\mu_p^T
\end{align}

Since 

\begin{equation}
    Var_{x,y} \begin{bmatrix} x/z \\ y/z \end{bmatrix} = E_{x,y}\left[ \begin{bmatrix} x/z \\ y/z \end{bmatrix}\begin{bmatrix} x/z \\ y/z \end{bmatrix}^T\right] - E_{x,y}\begin{bmatrix} x/z \\ y/z\end{bmatrix} E_{x,y}\begin{bmatrix} x/z \\ y/z\end{bmatrix}^T
\end{equation}

We get
\begin{equation}
    Var_{x,y}\begin{bmatrix} x/z \\ y/z \end{bmatrix} = \dfrac {\mu_z^2} {z^2} A^{-1}A^{-1} \dfrac {4+3\exp(-1/2)}{2+2\exp(-1/2)}
\end{equation}

\section{\OurDist}
\label{sec:supp_moments}

\subsection{Normalizing factor}
In this section we show
\begin{equation}
    K(\mu_z, A,a) = K_{depth}(\mu_z, a)K_{proj}(\mu_z, A)
\end{equation}

\textbf{proof}

\begin{align}
    K(\mu_z, A,a) &= \int\limits_{0}^{\infty} \int\limits_{-\infty}^{\infty}\int\limits_{-\infty}^{\infty} p_{proj}(x/z,y/z; \mu_z, A)dxdy p_{depth}(z; \mu_z, a) dz \\
    &= \int\limits_{0}^{\infty} K_{proj}(\mu_z, A) p_{depth}(z; \mu_z, a) dz \\
    &= K_{proj}(\mu_z, A) K_{depth}(\mu_z, a)
\end{align}

Since the normalizing factor of $p_{depth}$ does not depend on $z$

\subsection{Expected projected coordinates}
\begin{equation}
E\begin{bmatrix} x/z \\ y/z \end{bmatrix} = \mu_p
\end{equation}

\textbf{proof}
\begin{equation}
E_z\left[E_{x,y}\begin{bmatrix} x/z \\ y/z \end{bmatrix} \right ]= E_z\left[\mu_p\right] = \mu_p
\end{equation}

\subsection{Variance of projected coordinates}
The variance of the projected coordinates is
\begin{equation}
Var\begin{bmatrix} x/z \\ y/z \end{bmatrix} = A^{-2}\dfrac {4+3\exp(-1/2)} {2 + 2\exp(-1/2)}
\end{equation}

\textbf{proof}

When integrating this over z we get

\begin{align}
    E_{x,y,z}\begin{bmatrix} x/z \\ y/z \end{bmatrix} &= E_z\left[E_{x,y}\begin{bmatrix} x/z \\ y/z \end{bmatrix}\right] \\
    &= \mu_p\mu_p^T + \dfrac {4+3\exp(-1/2)} {2+2\exp(-1/2)} A^{-2}E_z\left[\mu_z^2/z^2 \right]
\end{align}

\begin{align}
    E\left[ \mu_z^2/z^2 \right] &= \dfrac {\mu_z^2} {K_{depth}(\mu_z, a)}
    \int_{0}^{\infty} \dfrac {1} {z^2} \exp(-a\exp(|\log(z/\mu_z)|)) dz \\
    &= \dfrac {\mu_z^2} {\mu(\Gamma(1,a)/a+\Gamma(-1,a)a)} \mu^{-1}(\Gamma(1,a)/a+\Gamma(-1,a)a) \\
    &= 1
\end{align}

Therefore
\begin{equation}
    Var\begin{bmatrix} x/z \\ y/z \end{bmatrix} = \dfrac {4+3\exp(-1/2)} {2+2\exp(-1/2)}A^{-2} + \mu_p\mu_p^T-\mu_p\mu_p^T
\end{equation}

\subsection{Expected depth}
The expected depth is
\begin{equation}
E\left[z\right] = \mu_z \dfrac {\Gamma(2, a)a^{-2} + \Gamma(-2, a)a^2} {\Gamma(1, a)a^{-1} + \Gamma(-1, a)a}
\end{equation}

\textbf{proof}

\begin{align}
E\left[z\right] &= \dfrac {1}{K_{proj}(A,\mu)K_{depth}(a,\mu_z)}\int\limits_{0}^{\infty} zp_{depth}(z; a, \mu_z)
\int\limits_{-\infty}^{\infty}
\int\limits_{-\infty}^{\infty} 
p_{proj}(x/z, y/z; A, \mu)dxdydz  \\
&= \dfrac {1}{K_{proj}(A,\mu)K_{depth}(a,\mu_z)}\int\limits_{0}^{\infty} zp_{depth}(z; a, \mu_z)
K_{proj}(A,\mu)dz = \mu_z \dfrac {\Gamma(2, a)a^{-2} + \Gamma(-2, a)a^2} {(\Gamma(1, a)a^{-1} + \Gamma(-1, a)a)^2}
\end{align}

\subsection{Depth variance}
\begin{equation}
Var\left[z\right] = \mu_z^2 \dfrac {(\Gamma(3,a)a^{-3} + \Gamma(-3, a)a^3)(\Gamma(1,a)/a+ \Gamma(-1, a)a)- (\Gamma(2,a)/a^2+ \Gamma(-2, a)a^2)^2} {(\Gamma(1,a)/a+ \Gamma(-1, a)a)^2}
\end{equation}

\textbf{proof}

\begin{equation}
E\left[z^2\right] = \E_z\left[\E_{x,y}\left[z^2\right]\right] = \E_z\left[z^2\right] = \mu_z^2 \dfrac {\Gamma(3,a)a^{-3} + \Gamma(-3, a)a^3} {\Gamma(1, a)/a+\Gamma(-1, a)a}
\end{equation}

and 
\begin{equation}
    Var[z] = E[z^2] - E[z]^2
\end{equation}

\section{Proof constraints}
\label{sec:supp_proof_constraints}
\subsection{Proof constraint 5}
\label{sec:supp_convexity}
We want to show that the negative log likelihood is convex with respect to $(x,y,z)$ for fixed $\mu, A, a$. 

Since the distribution decomposes into
\begin{equation}
    p_{combined}(x,y,z;\mu, A,a) = p_{proj}(x,y,z;\mu, A)p_{depth}(x,y,z;\mu, A)
\end{equation}

Therefore
\begin{equation}
    -\log(p_{combined}(x,y,z;\mu, A,a)) = -\log(p_{proj}(x,y,z;\mu, A))-\log(p_{depth}(x,y,z;\mu, A))
\end{equation}

Since convexity is preserved under addition it is sufficient to show that the negative log likelihood of $p_{proj}$ and $p_{depth}$ are convex separately.

\begin{equation}
-log(p_{proj}(x,y,z;\mu, A) = h\left(\left\|A\begin{bmatrix} x/z -\mu_x \\ y/z -\mu_x\end{bmatrix}\right\|_2\dfrac{z}{\mu_z}\right)
\end{equation}

Doing the affine parameter change 

\begin{align}
    q &= A\begin{bmatrix} x-z\mu_x \\ y-z\mu_y \end{bmatrix}/\mu_z \\
    u &= z/\mu_z
\end{align}

gives
\begin{equation}
    -log(p_{proj}(x,y,z;\mu, A) = h(\|q\|_2)
\end{equation}

which is convex.
Since convexity is preserved under affine transforms this concludes the proof for $p_{proj}$

The proof for $p_{depth}$ is even simpler
\begin{equation}
-\log(p_{depth}(z,\mu_z, a) = \max(az/\mu_z, a\mu_z/z)
\end{equation}

maximum of convex functions is convex. Therefore it is sufficient to show that each expression in the max is convex.

$az/\mu_z$ is linear, linear functions are convex

\begin{equation}
    \dfrac {\partial^2} {\partial z^2} a\mu_z/z = 2a\mu_z/z^3 > 0
\end{equation}
Therefore the second expression is convex as well.
This concludes the proof.
\subsection{Proof constraint 6}
\label{sec:supp_proof_constraint_6}
First consider the region $ z \le 0$. The statement is trivially true here since 0 converges to 0.

Secondly consider the region $0 < z <= \mu_z$

in this region $x_i^2+y_i^2 \rightarrow \infty$

\begin{align}
p(x,y,z) &= \dfrac {1}{K(A,\mu, a)}
\exp\left(-h\left(\left\|A\begin{bmatrix} x/z-\mu_x \\ y/z-\mu_y \end{bmatrix}\right\|_2\dfrac{z}{\mu_z}\right)\right)
\exp(-a \mu_z/z) \\
& \le \exp\left(-h\left(\left\|\dfrac {A} {\mu_z}\begin{bmatrix} x \\ y \end{bmatrix}\right\|_2 - \left\|A\begin{bmatrix} \mu_x \\ \mu_y \end{bmatrix}\right\|_2 \right)\right)
\dfrac {\exp(-a)} {K(A,\mu, a)}
\end{align}

The last step comes from applying the triangle inequality on the $\|\|_2$ term and maximizing the depth and projected terms independently.

For the last expression the first factor converges to 0 since $\mu_z$ is positive, $A$ is positive definite and $\|\mu_p\|_2$ is finite. The second factor is finite.
Therefore for limits in this region the condition holds.

In the region $\mu_z \le z< (x^2+y^2)^{1/4}$
also need $\sqrt{x^2+y^2}$ to go to infinity.

In this region the following inequalities hold. 
\begin{equation}
\max\limits_{\mu_z \le z< (x^2+y^2)^{1/4}} \exp(-az/\mu_z) \le e^{-a}
\end{equation}

Since the depth distribution is decreasing where the first step relaxes the constraint $\mu_z \le z< (x^2+y^2)^{1/4}$ to $\mu_z \le z$
and does the variable change $u = z/\mu_z$

The following inequality holds in this region as well.

\begin{align}
\min\limits_{ \mu_z<z<\sqrt{x^2+y^2}} 
\left\|\dfrac{A} {\mu_z}\begin{bmatrix}
x \\ y
\end{bmatrix} - \begin{bmatrix} \mu_x \\ \mu_y \end{bmatrix} \dfrac {z} {\mu_z} \right\|_2 &\ge
\min\limits_{\mu_z<z<\sqrt{x^2+y^2}} 
\left\|\dfrac{A} {\mu_z}\begin{bmatrix}
x \\ y
\end{bmatrix}\right\|_2 - \left\|\begin{bmatrix} \mu_x \\ \mu_y \end{bmatrix} \dfrac {z} {\mu_z} \right\|_2 \\
&\ge (x^2+y^2)^{1/4} \left(\dfrac {\lambda_{min}(x^2+y^2)^{1/4} - \|\mu_p\|} {\mu_z}\right)
\end{align}

which goes to infinity as $x^2+y^2$ goes to infinity.

Therefore
$p(x,y,z)$ goes to 0 in this region since
\begin{equation}
\exp\left(-h\left(\left\|A\begin{bmatrix} x/z-\mu_x \\ y/z-\mu_y \end{bmatrix}\right\|_2\dfrac{z}{\mu_z}\right)\right)
\end{equation}
goes to 0 while 
\begin{equation}
\dfrac {1}{K(A,\mu, a)} \exp(- az/\mu_z)
\end{equation}
is bounded by a constant.

For the region $(x^2+y^2)^{1/4} < z$ z will need to go to infinity.
the $\exp(-h())$ term is smaller than 1 and
\begin{equation}
\min\limits_{z \rightarrow \infty} \exp(-az/\mu_z) = 0
\end{equation}

The union of these regions is $\mathbb{R}^3$. The limit is also the same for each region, therefore the limit is the same for the union of the regions. which concludes the proof.
\subsection{Proof constraint 7}
\label{supp:proof_continous}
The function is continuous for all values when $z>0$ due to the fact that the density is a combination of continuous functions for this region.
The function is also continuous when $z\le0$ since the function is 0 in this region.
For the regions $\mu_z>z>0$ we know
\begin{align}
0 < p_{combined}(x,y,z) &= p_{proj}(x,y|z;\mu,A)p_{depth}(z;\mu_z,a) \le \dfrac{1}{K(\mu,A,a)} \exp(-a\mu_z/z)
\end{align}
The normalizing factor is finite and the exponential term goes to 0 as $z\rightarrow 0$


\section{Ground truth bounded after basis change}
\label{supp:proof_bound_gt_variables}
\begin{align}
z_p &= \dfrac{z} {f\mu_{z0}} \\
&= z/f /\sqrt{(\max\limits_{z, f \in Dataset}(z/f))(\min\limits_{z, f \in Dataset}(z/f))} \\
&\le \sqrt{\dfrac{\max\limits_{z, f \in Dataset}(z/f)}{\min\limits_{z, f \in Dataset}(z/f))}} = D
\end{align}

and 

\begin{align}
1/z_p &= \dfrac {f\mu_{z0}}{z} \\
&\le \dfrac {\sqrt{(\max\limits_{z, f \in Dataset}(z/f))(\min\limits_{z, f \in Dataset}(z/f))}} {(\min\limits_{z, f \in Dataset}(z/f))} \\
&= \sqrt{\dfrac{\max\limits_{z, f \in Dataset}(z/f)}{\min\limits_{z, f \in Dataset}(z/f))}} = D
\end{align}

$x_p$ and $y_p$ are bounded by 1 because
$(fx/z+S/2, fx/z+S/2)  \in \left[0, S\right] \times \left[0, S\right]$
for objects in the field of view of the camera.
Therefore
\begin{gather}
(fx/z, fy/z)  \in \left[-S/2, S/2\right] \times \left[-S/2, S/2\right] \Rightarrow \\
(2fx/Sz, 2fy/Sz)  \in \left[-1, 1\right] \times \left[-1, 1\right]
\end{gather}
Which concludes the proof

\section{Proof of bounded gradients for loss}
\label{supp:bound_grad}
This section contains a proof that the gradients are bounded with respect to the network output when the mapping in section \ref{subsec:parameter_remapping} is used.

The proof for $L_{proj}$ is already done in \cite{mohlin2021probabilistic}

The mapping from $w_1$ to $a$ is a contraction. Therefore it is sufficient to show that the loss has bounded gradients with respect to $a$ and $w_2$ to prove bounded gradients with respect to $w_1, w_2$

For the region $w_2 > 0$

the loss regression part of the loss is 
\begin{equation}
    \max((a+w_2)/z_p, z_p\dfrac {a^2}{a+w_2})
\end{equation}

The gradient of the first term is 
\begin{equation}
\|\nabla_{a, w_2} (a+w_2)/z_p\|_2 = \|(1/z_p,1/z_p)\|_2 \le \sqrt{2}D
\end{equation}

The gradient of the second expression is
\begin{equation}
\left\|\nabla_w z_p\dfrac {a^2}{a+w_2}\right\|_2 = 
z_p\left\|\begin{bmatrix} 
-\dfrac{a}{a+w_2}\dfrac{a+2w_2}{a+w_2}   \\
- \dfrac {a^2} {(a+w_2)^2}
\end{bmatrix}\right\|_2 \le D\sqrt{5}
\end{equation}

If $w_2 \le 0$
the loss is
\begin{equation}
\max\left(z_p(a-w_2), \dfrac{1}{z_p} \dfrac{a^2}{a-w_2}\right)
\end{equation}

The first expression has gradient norm

\begin{equation}
\|\nabla_{a, w_2} z_p(a-w_2)\|_2 = z_p\|(1,-1)\|_2 \le \sqrt{2}D
\end{equation}

The second expression has gradient norm

\begin{equation}
\left\|\nabla_{a,w_2} \dfrac{1}{z_p} \dfrac {a^2}{a-w_2}\right\|_2 = 
\dfrac{1} {z_p} \left\|\begin{bmatrix} 
-\dfrac{a}{a-w_2}\dfrac{a-2w_2}{a-w_2}   \\
\dfrac {a^2} {(a-w_2)^2}
\end{bmatrix}\right\|_2 \le D\sqrt{5}
\end{equation}

Which concludes for the regression part of the loss.

The normalizing factor component of the depth loss is

\begin{equation}
\log(\exp(-a)/a +\Gamma(-1,a)a)
\end{equation}

for The gradient of this is 

\begin{equation}
    \dfrac {\partial \log(\Gamma(1,a)/a+\Gamma(-1,a)a)} {\partial a} = -\dfrac{1}{a} + 2\dfrac {1+1/a} {1+a^2\Gamma(-1,a)\exp(a)}
\end{equation}

Which is less than $4$ when $a>1$

If $a < 1$ the mapping between $w_2$ and $a$ is $a(w_1) = exp(w_1)$
Therefore

\begin{align}
    \dfrac {\partial \log(\Gamma(1,a(w_1))/a(w_1)+\Gamma(-1,a(w_1))a(w_1))} {\partial w_1} &= (-\dfrac{1}{a} + \dfrac {1+1/a} {1+a^2\Gamma(-1,a)\exp(a)})\dfrac {\partial a}{\partial w_1} \\
    &= (-1 + 2\dfrac {a(w_1)+1} {1+a(w_1)^2\Gamma(-1,a(w_1))\exp(a(w_1))})
\end{align}

Which is less than $4$. Therefore the negative log likelihood of the normalizing factor has bounded gradients with respect to $w_1$

\section{Computing normalizing factor}
\label{sec:supp_normalizing_factor}
The normalizing factor can be rewritten as 

\begin{equation}
    log(\Gamma(1,a)/a+\Gamma(-1,a)a) = -log(a)-a+\log(1+\Gamma(-1,a)a^2\exp(a))
\end{equation}

with gradient

\begin{equation}
    \dfrac {\partial log(\Gamma(1,a)/a+\Gamma(-1,a)a)} {\partial a} = -1/a + 2\dfrac{1+1/a}{1+a^2\Gamma(-1,a)\exp(a)}
\end{equation}

Therefore if $a^2\Gamma(-1,a)\exp(a)$ can be computed accurately both the function and its gradient can be computed. We show how to do this in section \ref{sec:supp_numerical_integration_section}

\textbf{Proof of function evaluation}
\begin{align}
\log(\Gamma(1,a)/a+\Gamma(-1,a)a)
&= \log(\exp(-a)/a+\Gamma(-1,a)a) \\ 
&= \log(\exp(-a)/a(1+\Gamma(-1,a)a^2\exp(a))) \\
&= \log(1+\Gamma(-1,a)a^2\exp(a))-a-\log(a)
\end{align}

\textbf{Proof of gradient}
\begin{align}
\dfrac {\partial log(1+\Gamma(-1,a)a^2\exp(a))-a-\log(a)} {\partial a}
&= -1-1/a+\dfrac{(2a+a^2)\Gamma(-1,a)\exp(a)-a^2\exp(a)\dfrac{\partial\Gamma(-1,a)}{\partial a}} {1+\Gamma(-1,a)a^2\exp(a)} \\
&= -1 -1/a + \dfrac{(2/a+1)a^2\Gamma(-1,a)\exp(a) - 1} {1+\Gamma(-1,a)a^2\exp(a)} \\
&= -1 -1/a + \dfrac{(2/a+1)(a^2\Gamma(-1,a)\exp(a) +1) -(2/a+1) - 1} {1+\Gamma(-1,a)a^2\exp(a)} \\
&= -1 -1/a (1+2/a) + \dfrac{-(2/a+2)} {1+\Gamma(-1,a)a^2\exp(a)} \\
&= 1/a - 2\dfrac {1/a+1} {1+\Gamma(-1,a)a^2\exp(a)}
\end{align}

\subsection{numerical integration of $a^2\Gamma(-1,a)
\exp(a)$}
\label{sec:supp_numerical_integration_section}
Here we show that 

\begin{equation}
    a^2\Gamma(-1,a)\exp(a) = \int\limits_{0}^{\infty} \dfrac {1} {(1+\log(1/y)/a)^2} dy
\end{equation}

\textbf{proof}

\begin{align}
    a^2\Gamma(-1,a)\exp(a) &= \int\limits_{a}^{\infty}\dfrac {a^2} {t^2} \exp(a-t) dt \\
    &= \int\limits_{0}^{1}\dfrac {1} {(\log(1/y)/a+1)^2}) dy
\end{align}

where that basis change $y=exp(-t+a)$ is used.

\section{Convex loss with respect to network output}
\label{supp:convex_parameters}
Here we show that the network is convex with respect  to the network outputs

\textbf{Proof}
In the following sections we will prove that both the projected and depth loss are convex with repect to their parameters. To prove the depth loss is convex.

\subsection{Projected term}
We have the same loss and the same mapping to parameterize $B$ and $\nu_p$ as in \cite{mohlin2021probabilistic}
In this work they prove that this loss is convex when $B\succ \theta$. We use $\theta = 1$ which concludes the proof.

\subsection{Depth Loss}

\subsubsection{Rewriting argument of logarithm of depth normalizing factor}
\label{sec:supp_convex_depth_loss_rewrite_integrand}

In this section we show that

\begin{equation}
\label{eq:cov_proff_integrand_log_arg}
    1/a + a\Gamma(-1, a)\exp(a) = \int\limits_{0}^{1} \dfrac {2a^2-2a\log(1/y)+\log^2(1/y)} {a(a+\log(1/y))^2} dy
\end{equation}

\textbf{proof}

\begin{align}
    1/a + a\Gamma(-1, a)\exp(a) &= \int\limits_{0}^{1} \dfrac {1}{a} + \dfrac{1}{a}\dfrac {1}{(1+\log(1/y)/a)^2} dy \\
    &= \int\limits_{0}^{1} \dfrac {1}{a} + \dfrac {a}{(a+\log(1/y))^2} dy \\
    &= \int\limits_{0}^{1} \dfrac {a^2+2a\log(1/y)+\log^2(1/y)+ a^2}{a(a+\log(1/y))^2} dy
\end{align}

\subsubsection{Log-convexity of integrand in equation \ref{eq:cov_proff_integrand_log_arg}}

\label{sec:supp_convex_log_convex_integrand}

Here we show that for $a>0$ and $y \in [0,1)$

\begin{equation}
q_{integrand}(y,a) = \dfrac {a^2+2a\log(1/y)+\log^2(1/y)+ a^2}{a(a+\log(1/y))^2}
\end{equation}

is log-convex with respect to $a$

\textbf{proof}

since $log(1/y)$ is a constant for this proof we denote this value as $k$
\begin{align}
\dfrac {\partial^2} {\partial a^2} \log(\dfrac {2a^2+2ak+k^2}{a(a+k)^2} &= \dfrac {\partial^2} {\partial a^2} \log(2a^2+2ak+k^2) - \log(a) -2\log(a+k) \\
&= \dfrac {\partial} {\partial a} \dfrac {4a+2k} {2a^2+2ak+k^2} - 1/a -2/(a+k) \\
&=\dfrac {4(a^2+(a+k)^2)+(4a+2k)^2} {(a^2+(a+k)^2)^2} + 1/a^2 + 2/(a+k)^2
\end{align}

To show log convexity we need to show that this expression is larger than 0. To make notation slighly easier we denote $t = a+k$, since both a and k are positive t is positve as well.

\begin{align}
    \dfrac {\partial^2} {\partial a^2} \log(\dfrac {2a^2+2ak+k^2}{(a(a+k)^2})
    &= \dfrac {4(a^2+t^2)-(2a+2t)^2} {(a^2+t^2)^2} + 1/a^2 + 2/t^2 \\
    &= \dfrac {-8a^3t^3 + (t^2+2a^2)(a^2+t^2)^2} {(a^2+t^2)^2} \\
    &= \dfrac {-8a^3t^3 + (t^2+2a^2)(a^2+t^2)^2} {(a^2+t^2)^2} \\
    &= \dfrac {t^6+4t^4a^2-8a^3t^3+5t^2a^4 +2a^6} {(a^2+t^2)^2} \\
    &=\dfrac {t^6+4t^2a^2(a-t)^2
    t^2a^4 +2a^6} {(a^2+t^2)^2} \ge 0
\end{align}

In the last step every term is positive in both numerator and denominator.

\subsubsection{log-convex integral for log-convex integrands}
\label{sec:supp_convex_integrands_to_integral}
For a function
\begin{equation}
    g(a) = \int f(x,a) dx
\end{equation}
where $f$ is continuous and log-convex with respect to a, then g(a) is log-convex as well.

\textbf{proof}
First we show
\begin{equation}
    \log(g(a))/2 + \log(g(b))/2 \ge \log(g((a+b)/2))
\end{equation}

which is equivalent to 
\begin{equation}
g(a)g(b) \ge (g((a+b)/2))^2
\end{equation}
Since $\exp$ is increasing

\begin{align}
g(a)g(b) &= \int f(x,a) dx \int f(x,b)dx  \\
&\le \left(\int \sqrt{f(x,a)} \sqrt{f(x,b)}dx\right)^2\\
&= \left(\int \exp(1/2\log(f(x,a))+1/2\log(f(x,b)))dx\right)^2\\
&\le \left(\int \exp(\log(f(x,(a+b)/2)))dx\right)^2\\
&= \left(\int (f(x,(a+b)/2))\right)^2 = (g((a+b)/2))^2
\end{align}

The first inequality is cauchy-schwartz, the second comes from the fact that $f$ is log convex.

The proof can be extended for all rational $\theta$ between 0 and 1 where the denominator is a power of 2 by recursive bisection. 
\begin{equation}
    log(g(a))\theta + log(g(a))(1-\theta) \ge \log(g(a\theta+(1-\theta)b))
\end{equation}
Using continuity completes the proof for all $\theta \in [0,1]$

Which concludes the proof.

\subsubsection{convexity of $\log(\Gamma(1,a)/a+\Gamma(-1,a)a)$}
In this section we show that
\begin{equation}
    \log(\Gamma(1,a)/a+\Gamma(-1,a)a)
\end{equation}
is convex with respect to $a$

\textbf{proof}

This is the same as
\begin{equation}
    1/a+\Gamma(-1,a)a\exp(a)
\end{equation}
being log convex.

Now the proof follows from section \ref{sec:supp_convex_depth_loss_rewrite_integrand}, \ref{sec:supp_convex_log_convex_integrand} and \ref{sec:supp_convex_integrands_to_integral}

\subsubsection{depth loss error term}
here we show that 
\begin{equation}
a\max(z/\mu, \mu/z)
\end{equation}
is convex when 
\begin{equation}
    \mu = \begin{cases}
         u/a+1 &\text{ if }  u > 0 \\
         1/(1-u/a) &\text{ otherwise}
    \end{cases}
\end{equation}

\textbf{proof}
if $u \ge 0$ then
\begin{align}
\mu/z &= (u+a)/z \\
az/\mu &= za^2/(u+a)
\end{align}
The first expression is linear therefore convex.

The hessian of the second expression is
\begin{align}
    \dfrac {\partial^2 } {\partial a^2} za^2/(u+a) \\
    &= \dfrac {\partial } {\partial a} z(2ua+a^2)/(u+a)^2 \\
    &= z(2(u+a)(u+a)-2(2ua+a^2))/(u+a)^3 \\
    &= 2z(u^2+2au+u^2-2ua+a^2)/(u+a)^3 \\
    &= 2zu^2/(u+a)^3 \\
    \dfrac {\partial^2 } {\partial a\partial u} za^2/(u+a)
    &= z(2a(a+b)-2(2ua+a^2))/(u+a)^3 \\
    &= -2zua/(u+a)^3 \\
    \dfrac {\partial^2 } {\partial u} za^2/(u+a) 
    &= \dfrac {\partial } {\partial u} -za^2/(u+a) \\
    &= -2za^2/(u+a)^2
\end{align}

which has the eigenvalues 0 and $4z(a^2+u^2)/(a+u)^3 > 0$. The denominator is positive since both $a$ and $u$ are positive.
Therefore this expression is positive definite.

If $u < 0$ Then the expression 
$az/\mu = a-u$ which is linear and therefore convex.
The expression
\begin{equation}
    a\mu/z = a^2/(a-u)
\end{equation}
doing the variable change $v = -u$ turns this in the expression
\begin{equation}
    a\mu/z = a^2/(a+v)
\end{equation}
which we have already shown is convex for $v > 0$.
Therefore this expression is convex for this region.
max of convex expressions is convex therefore the function is convex in each region.
The mapping between $u$ and $\mu$ has continous gradient at the boundary $u=0$ Therefore the function is convex for all $a>0, u\in\mathbb{R}$. Since mapping between network output and $a$ is linear for the region $a>1$ this the loss is convex for this region.

This concludes the proof that the loss is convex in the region $A \succ 1, a > 1$

\end{document}